\documentclass{article}
\usepackage{iclr2027_conference,times}

\usepackage{amsmath,amsfonts,bm}

\def\eqref#1{equation~\ref{#1}}

\def\1{\bm{1}}

\DeclareMathAlphabet{\mathsfit}{\encodingdefault}{\sfdefault}{m}{sl}
\SetMathAlphabet{\mathsfit}{bold}{\encodingdefault}{\sfdefault}{bx}{n}

\usepackage{graphicx}
\usepackage{multirow}
\usepackage{booktabs}
\usepackage{enumitem}
\usepackage{wrapfig}
\usepackage{hyperref}
\usepackage{url}
\AddToHook{env/table/begin}{\setlength{\abovecaptionskip}{0pt}\setlength{\belowcaptionskip}{6pt}}
\AddToHook{env/table*/begin}{\setlength{\abovecaptionskip}{0pt}\setlength{\belowcaptionskip}{6pt}}
\hypersetup{
  pdftitle={ReCast: Attribution-Oriented Step Representation Learning for LLM-Based Agent Systems},
  pdfauthor={Weilin Jin, Mingyu Wang, Taiyu Zhu, Ziqi Zhou, Wenbo Li, Haoyang Huang, Nan Duan, Yifan Wu, Ying Li, Zhonghai Wu},
  colorlinks=true,
  linkcolor=blue,
  citecolor=blue,
  urlcolor=blue
}

\title{ReCast: Attribution-Oriented Step Representation Learning for LLM-Based Agent Systems}

\author{%
    Weilin Jin\textsuperscript{1,2},
    Mingyu Wang\textsuperscript{1,2},
    Taiyu Zhu\textsuperscript{1},
    Ziqi Zhou\textsuperscript{3},
    Wenbo Li\textsuperscript{2}\thanks{Project leader.}\,,\\
    \bfseries Haoyang Huang\textsuperscript{2},
    Nan Duan\textsuperscript{2},
    Yifan Wu\textsuperscript{1}\thanks{Corresponding authors.}\,,
    Ying Li\textsuperscript{1}\footnotemark[2]\,,
    Zhonghai Wu\textsuperscript{1}\\[0.5em]
    \textnormal{\textsuperscript{1}Peking University, Beijing, China}\\
    \textnormal{\textsuperscript{2}Joy Future Academy, Beijing, China}\\
    \textnormal{\textsuperscript{3}Tsinghua University, Beijing, China}
}

\iclrfinalcopy

\begin{document}

\maketitle
\lhead{}

\begin{abstract}
% !TEX root = ../main.tex
In LLM-based agent systems, failures can originate from early steps whose effects propagate through subsequent interactions, making their origins difficult to identify. To trace such failures back to their origin, failure attribution has been formulated as the task of identifying the earliest step responsible for the failure. Recent methods leverage LLM internal signals for failure attribution, typically using hidden states as step representations. We therefore conduct an empirical study to evaluate how effectively these representations distinguish root-cause steps from other steps and find limited separation. Motivated by this observation, we propose \textbf{ReCast}, a step representation learning method that transforms hidden states from a frozen LLM into attribution-oriented step representations. ReCast first selects attribution-relevant layers, then constructs complementary pattern and deviation features, and finally learns contextualized step representations through an encoder trained with contrastive and ranking objectives. We also introduce ReCast-2K, a training dataset for failure attribution. ReCast achieves the best Hit@1 across four benchmarks, surpassing the strongest baseline by 5.65 and 9.19 pp on Who\&When Algorithm and Handcrafted, respectively. Code is available at {\urlstyle{same}\url{https://anonymous.4open.science/r/ReCast-5FB6}}.

\end{abstract}

% !TEX root = ../main.tex
\section{Introduction}

Large language model (LLM) based multi-agent systems (MASs) coordinate specialized agents, external tools, and environment interactions to solve increasingly complex tasks~\citep{wu2024autogen,fourney2024magentic,song2025captainagent}. Each execution produces a trace of agent activity, such as messages and tool calls, providing a basis for understanding system behavior. Despite their growing capabilities, MASs remain vulnerable to errors such as reasoning errors and incorrect tool use~\citep{liu2026whowhenpro}. These errors can propagate through subsequent interactions and ultimately cause task failure, limiting the reliability of MASs in real-world applications.

Diagnosing why a multi-agent execution failed is considerably harder than detecting the failure itself. A task may admit many valid solution paths, and stochastic agent decisions can produce substantially different traces even under the same system and task. More importantly, the root cause may initially appear plausible, since its consequences emerge only after several downstream steps. Failure attribution therefore requires identifying the earliest decisive error from a long sequence of actions, rather than merely detecting anomalous content or locating the final visible symptom~\citep{zhang2025whowhen}. Accurate step-level attribution is essential for debugging agent behaviors, improving orchestration strategies, and constructing more reliable multi-agent systems.

\begin{figure}[t]
\centering
\includegraphics[width=\textwidth]{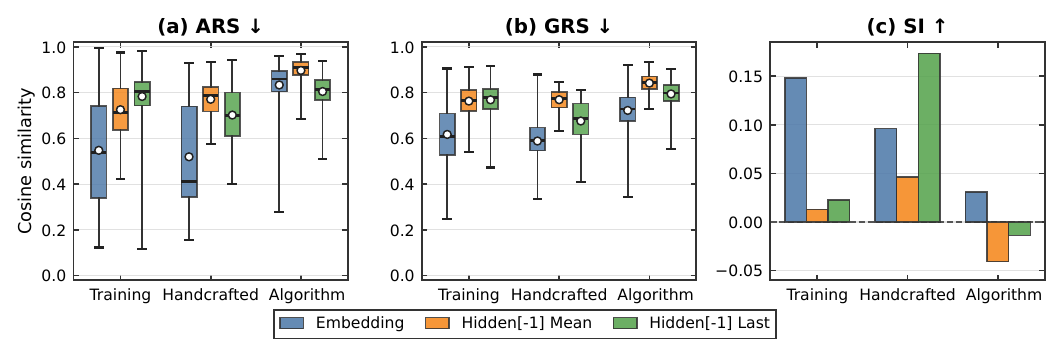}
\caption{Comparison of text embeddings and last-layer hidden-state representations on failed trajectories from ReCast-2K and the Handcrafted and Algorithm subsets of Who\&When. Lower ARS/GRS and higher SI indicate better separation. See Appendix~\ref{app:empirical-similarity-study} for metric definitions.}
\label{fig:observation-results}
\end{figure}

Existing methods such as CHIEF~\citep{wang2026chief} use LLMs to analyze execution traces, but autoregressive decoding can incur substantial inference latency. To reduce this overhead, recent methods explore failure attribution using LLM internal signals: MASPrism~\citep{liu2026masprism} derives negative log-likelihood from hidden states and combines it with attention. StepFinder~\citep{zhu2026stepfinder} instead learns attribution from text embeddings. These approaches involve different choices about how to represent each step for attribution. However, a key question remains underexplored: \emph{what step representations are most suitable for failure attribution?} To investigate this question, we empirically assess how well text embeddings and hidden states distinguish the root-cause step from other steps. Figure~\ref{fig:observation-results} reveals limited separation in these representations, motivating the construction and learning of attribution-oriented step representations.

To address this gap, we propose \textbf{ReCast}, a step representation learning method that integrates feature selection, feature construction, and contextual learning for failure attribution. The method consists of three modules. First, an attribution-aware probe selects internal features according to their ability to distinguish the root-cause step from other steps. Second, a fixed sparse Johnson--Lindenstrauss projection~\citep{li2006very} transforms the selected features into complementary pattern and deviation representations, capturing each step's internal activation structure and its departure from successful execution behavior. Third, a bidirectional encoder learns contextualized step representations from the complete trajectory. Training jointly optimizes root and non-root contrastive losses alongside a trajectory-level ranking loss. Together, these three stages determine which internal signals are retained, how they are represented, and how they are learned for attribution.

We also introduce \textbf{ReCast-2K}, a training dataset containing $1{,}044$ successful and $1{,}496$ failed trajectories. Experiments on four benchmarks---Who\&When~\citep{zhang2025whowhen}, TraceElephant~\citep{chen2026traceElephant}, AFTraj-2K~\citep{zhang2026agentforesight}, and Who\&When Pro~\citep{liu2026whowhenpro}---show that ReCast consistently achieves the best Hit@1, surpassing the strongest baseline by 5.65 and 9.19 pp on Who\&When Algorithm and Handcrafted, respectively. Ablation studies further support the contributions of its main components. Our main contributions are as follows:
\begin{itemize}[leftmargin=1.2em]
\item To the best of our knowledge, we are the first to systematically study the discriminative ability of step representations for multi-agent failure attribution, using quantitative metrics to assess separation within and across trajectories.
\item We propose ReCast, a step representation learning method that combines feature selection, feature construction, and contextual learning to produce attribution-oriented step representations.
\item We construct ReCast-2K, a diverse source training dataset, and evaluate ReCast on four benchmarks, where it outperforms all compared baselines in Hit@1 accuracy. ReCast also enables efficient attribution from precomputed features.
\end{itemize}

% !TEX root = ../main.tex
\section{Related Work}

\textbf{Black-box Failure Attribution Methods.} These methods attribute failures from observable trajectories through structured reasoning or learned models. Among training-free methods, CHIEF~\citep{wang2026chief} constructs hierarchical causal graphs from execution logs for oracle-guided backtracking and counterfactual attribution. FALAT~\citep{rafi2026falat} formulates attribution as hierarchical dependency-guided search. Distinguishing root causes from noisy or propagated symptoms remains challenging for these methods.

Learning-based methods train attribution models from trajectories. AgenTracer~\citep{zhang2026agentracer} generates annotated failures through counterfactual replay and fault injection and trains AgenTracer-8B with reinforcement learning. AgentForesight~\citep{zhang2026agentforesight} detects decisive errors from trajectory prefixes for online auditing. StepFinder~\citep{zhu2026stepfinder} models cross-step dependencies in semantic sequences for root-cause ranking. ASCon~\citep{jiang2026ascon} jointly models step context and agent interactions for failure attribution. These methods primarily use external trajectory representations, leaving internal diagnostic signals underexplored.

\textbf{White-box Failure Attribution Methods.} These methods exploit LLM internal signals. MASPrism~\citep{liu2026masprism} identifies symptom steps using negative log-likelihood (NLL) derived from prefill-stage hidden states, then traces potential root causes using attention weights. OAT~\citep{yeh2026oat} learns successful hidden-state dynamics with neural controlled differential equations and attributes failures to steps that deviate from these dynamics. Rather than directly scoring internal signals or modeling successful dynamics, ReCast learns attribution-oriented step representations.

% !TEX root = ../main.tex
\section{PROBLEM FORMULATION}

\subsection{Definition}
\label{sec:problem-definition}

\textbf{Background.} Consider a multi-agent system with an agent set $\mathcal{M}=\{m_1,\ldots,m_N\}$ interacting with an environment $\mathcal{E}$. Given a task query $q\in\mathcal{Q}$, the system produces an execution trace
\begin{equation}
\label{eq:multi-agent-trace}
\tau=(q, s_0, a_1, s_1, a_2, \ldots, a_t, s_t, \ldots, a_T, s_T),
\end{equation}
where $s_t$ is the system state after step $t$, $a_t$ is the candidate model action, and $T$ is the final step of the execution. Here, we only consider model actions as candidates for failure attribution. Let $B$ index a batch of traces, with $\tau_i$ denoting the trace indexed by $i\in B$ and $T_i$ its terminal step. Let $Z(\tau)\in\{0,1\}$ denote the task-specific success evaluator for task $q$, where $Z(\tau)=1$ indicates successful completion. We then define the successful and failed trace sets as $\mathcal{T}^{\mathrm{succ}}=\{\tau\mid Z(\tau)=1\}$ and $\mathcal{T}^{\mathrm{fail}}=\{\tau\mid Z(\tau)=0\}$.

\textbf{Objective.} Following Who\&When~\citep{zhang2025whowhen}, consider a failed trace $\tau_i$ with $Z(\tau_i)=0$. Suppose that the action $a_t$ taken by an active agent at step $t$ is replaced with a feasible corrective action $\tilde{a}_t$, while the execution history before step $t$ remains fixed. The system then re-executes the trajectory from step $t$ under the corrected action, producing a counterfactual trace $\tilde{\tau}_i$. If $Z(\tilde{\tau}_i)=1$, timestamp $t$ is defined as a \emph{decisive error}. We use the earliest decisive error as the unique \emph{root-cause} target for each failed trace $\tau_i$:
\begin{equation}
\label{eq:root-cause}
t_i^*=\min\{t\in\{1,\ldots,T_i\}:t\text{ is a decisive error in }\tau_i\}.
\end{equation}
The root-cause timestamp $t_i^*$ is unique by definition. We call all remaining candidate timestamps \emph{non-root} steps and write $\mathcal{N}(\tau_i)=\{1,\ldots,T_i\}\setminus\{t_i^*\}$. This label identifies steps other than the attribution target and does not imply that they are error-free. A successful trace has no root cause, so all its candidate timestamps belong to $\mathcal{N}(\tau_i)$. Failure attribution ranks candidate timestamps to identify the unique target $t_i^*$.

\subsection{Observation}
\label{sec:observation}

Some existing failure attribution methods use step representations derived from text embeddings or LLM hidden states. We conduct an empirical study to evaluate how effectively these representations distinguish the root-cause step from other steps. Specifically, we examine Qwen3 text embeddings~\citep{zhang2025qwen3embedding}, following StepFinder~\citep{zhu2026stepfinder}, and last-layer hidden states from a frozen LLM, motivated by MASPrism's use of internal signals~\citep{liu2026masprism}.

We assess these representations from two perspectives: \textbf{(1) Within-trace separation.} The root-cause step should be distinguishable from non-root-cause steps within the same trace. \textbf{(2) Cross-trace separation.} Root-cause representations should also be distinguishable from non-root representations across traces. Based on these criteria, we define three metrics. Adjacent Root Similarity (ARS) and Global Root Similarity (GRS) measure local and overall within-trace separation, respectively, while the cosine Silhouette Index (SI) measures cross-trace separation. Using these metrics, we compare Qwen3 text embeddings with mean-pooled and last-token representations from the final hidden-state layer on the failed trajectories in our training dataset and the Handcrafted and Algorithm subsets of Who\&When. See Appendix~\ref{app:empirical-similarity-study} for metric and representation details.

Figure~\ref{fig:observation-results} shows that the evaluated representations exhibit limited separation under the proposed metrics. High ARS and GRS indicate that the root-cause step remains similar to its predecessor and other steps in the same trace, while low SI indicates weak separation across traces. Although their relative performance varies, these representations exhibit limited separation between the root-cause step and other steps under these metrics. This observation motivates investigating whether attribution-oriented step representation learning can improve failure attribution.

% !TEX root = ../main.tex
\section{METHODOLOGY}

\begin{figure}[t]
\centering
\includegraphics[width=\textwidth]{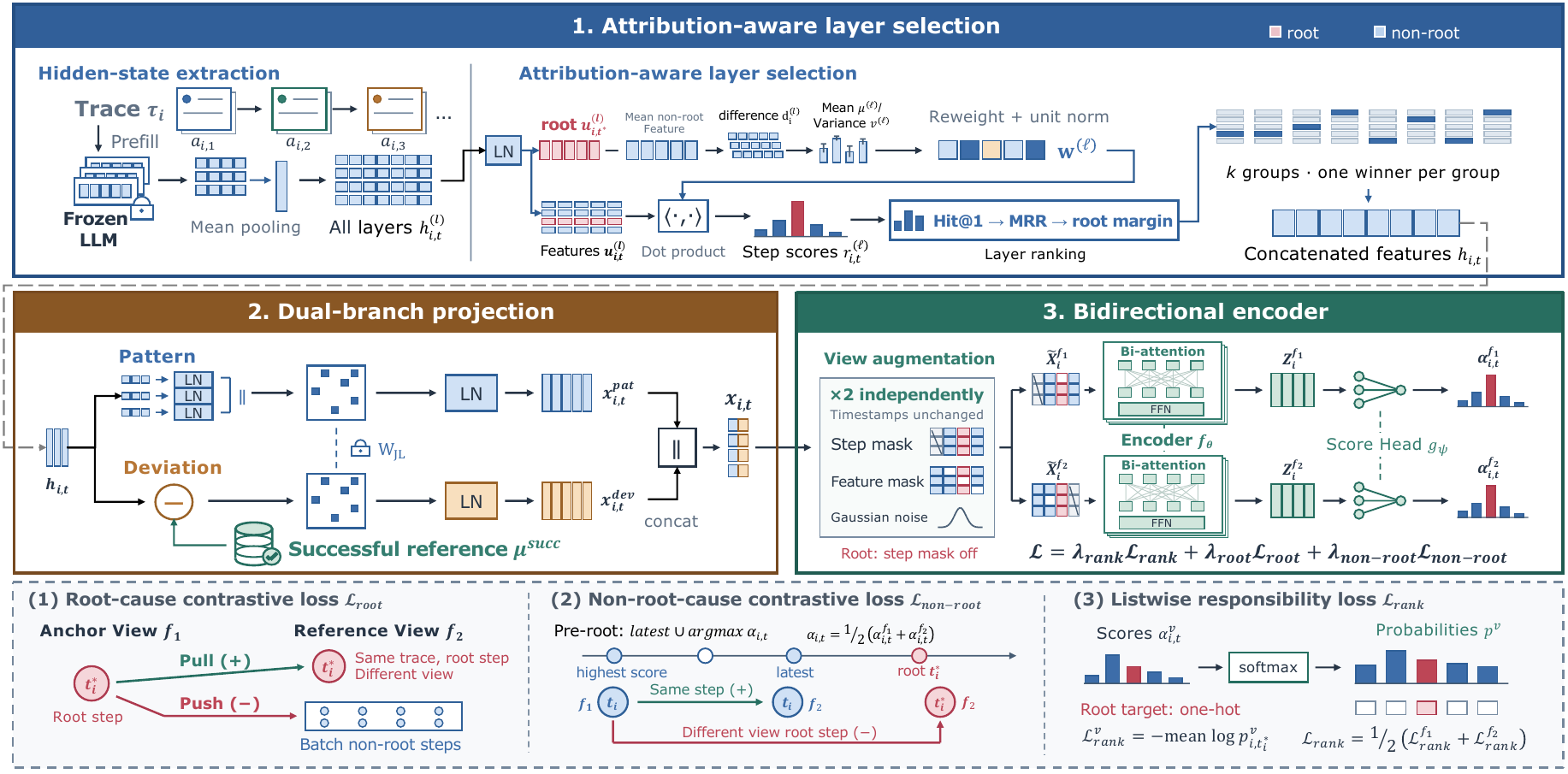}
\caption{Overview of ReCast, which learns attribution-oriented step representations from frozen LLM hidden states for failure attribution.}
\label{fig:method-overview}
\end{figure}

Motivated by our empirical study, we develop ReCast to learn attribution-oriented step representations from hidden states extracted by prefilling trajectories with a frozen LLM. As shown in Figure~\ref{fig:method-overview}, ReCast selects attribution-relevant layers and constructs complementary pattern and deviation features. An encoder is then jointly trained with a scoring head under root and non-root contrastive losses and a trajectory-level ranking loss to learn contextualized step representations.

\subsection{Feature Selection and Construction}
\label{sec:input-representation}

\textbf{Hidden-state extraction.} We use a frozen LLM with hidden-state dimension $d_h$ to extract step representations by prefilling the trajectory. For each model-generated action $a_{i,t}$ in trajectory $\tau_i$, we obtain the layer-$\ell$ step feature by mean-pooling its $d_h$-dimensional token-level hidden states, conditioned on the preceding execution prefix:
\begin{equation}
\label{eq:step-hidden-state}
\mathbf{h}_{i,t}^{(\ell)}
=
\mathrm{MeanPool}_{\mathrm{tokens}(a_{i,t})}
\left(
 \mathrm{LLM}^{(\ell)}(a_{i,t}\mid\tau_{i,<t})
\right),\qquad \mathbf{h}_{i,t}^{(\ell)}\in\mathbb{R}^{d_h}.
\end{equation}

\textbf{Attribution-aware layer selection.} Transformer layers encode information at different levels of abstraction and are not equally informative for failure attribution. We therefore use a lightweight probe to select layers according to their ability to distinguish the root-cause step from all other steps.
For each layer, we apply layer normalization (LN) to obtain normalized step features $\mathbf{u}_{i,t}^{(\ell)}$. We then construct a difference vector for each failed trajectory by subtracting the mean non-root feature from the root-cause feature:
\begin{equation}
\label{eq:layer-probe-contrast}
\mathbf{u}_{i,t}^{(\ell)}
=
\mathrm{LN}\left(\mathbf{h}_{i,t}^{(\ell)}\right),
\qquad
\mathbf{d}_{i}^{(\ell)}
=
\mathbf{u}_{i,t_i^*}^{(\ell)}
-
\frac{1}{|\mathcal{N}(\tau_i)|}
\sum_{t\in\mathcal{N}(\tau_i)}
\mathbf{u}_{i,t}^{(\ell)},\qquad \mathbf{u}_{i,t}^{(\ell)},\mathbf{d}_{i}^{(\ell)}\in\mathbb{R}^{d_h}.
\end{equation}
Using the difference vectors $\mathbf{d}_{i}^{(\ell)}$, we assess each layer's attribution capability through five-fold cross-validation on failed training trajectories. In each round, we compute $\boldsymbol{\mu}^{(\ell)},\mathbf{v}^{(\ell)}\in\mathbb{R}^{d_h}$ as the mean and variance vectors of $\mathbf{d}_{i}^{(\ell)}$ over the trajectories in four folds. These statistics define a scoring direction that emphasizes large and consistent differences:
\begin{equation}
\label{eq:layer-probe-direction-main}
\mathbf{w}^{(\ell)}
=
\frac{\boldsymbol{\mu}^{(\ell)}/(\mathbf{v}^{(\ell)}+\eta^{(\ell)})}
{\left\|\boldsymbol{\mu}^{(\ell)}/(\mathbf{v}^{(\ell)}+\eta^{(\ell)})\right\|_2},\qquad \mathbf{w}^{(\ell)}\in\mathbb{R}^{d_h}.
\end{equation}
Here $\eta^{(\ell)}=\lambda_{\mathrm{ridge}}d_h^{-1}\sum_{j=1}^{d_h}v_{j}^{(\ell)}$ is a scalar ridge term. We then evaluate this direction on the remaining fold by scoring each candidate step as $r_{i,t}^{(\ell)}=\langle\mathbf{u}_{i,t}^{(\ell)},\mathbf{w}^{(\ell)}\rangle$. We aggregate the evaluation results across all five folds and rank layers by their root-cause identification performance. Detailed selection criteria are provided in Appendix~\ref{app:method-layer-selection}.

Using this ranking, we select $k$ layers while ensuring coverage across the model depth. We partition all layers into $k$ contiguous groups and retain the highest-ranked layer from each group. Let $\ell_1,\ldots,\ell_k$ denote these layers in depth order; their step features are concatenated as
\begin{equation}
\label{eq:layer-concat}
\mathbf{h}_{i,t}
=
\left[
\mathbf{h}_{i,t}^{(\ell_1)};
\ldots;
\mathbf{h}_{i,t}^{(\ell_k)}
\right],\qquad \mathbf{h}_{i,t}\in\mathbb{R}^{kd_h}.
\end{equation}

\textbf{Dual-branch projection.} To reduce the dimensionality of the concatenated hidden state, we construct two complementary branches using a fixed sparse Johnson--Lindenstrauss projection~\citep{li2006very}. The pattern branch applies layer-wise normalization to preserve relative activation patterns. For the deviation branch, we introduce successful trajectories as a reference to capture how each step differs from successful execution. The branch outputs are computed as
\begin{equation}
\label{eq:dual-jl-output}
\mathbf{x}_{i,t}^{\mathrm{pat}}
\!=\!
\mathrm{LN}\!\left(\mathbf{W}_{\mathrm{JL}}\left[\mathrm{LN}\!\left(\mathbf{h}_{i,t}^{(\ell_j)}\right)\right]_{j=1}^{k}\right)\in\mathbb{R}^{d_p},
\qquad
\mathbf{x}_{i,t}^{\mathrm{dev}}
\!=\!
\mathrm{LN}\!\left(\mathbf{W}_{\mathrm{JL}}\left(\mathbf{h}_{i,t}-\boldsymbol{\mu}^{\mathrm{succ}}\right)\right)\in\mathbb{R}^{d_p}.
\end{equation}
Here, $\boldsymbol{\mu}^{\mathrm{succ}}\in\mathbb{R}^{kd_h}$ is the mean concatenated hidden state over all steps in successful training trajectories, and $\mathbf{W}_{\mathrm{JL}}\in\mathbb{R}^{d_p\times kd_h}$ is the sparse projection matrix shared by both branches. In the pattern branch, normalization is applied separately to each selected layer before concatenation; the deviation branch operates directly on the concatenated hidden state. The outputs are concatenated as $\mathbf{x}_{i,t}=[\mathbf{x}_{i,t}^{\mathrm{pat}};\mathbf{x}_{i,t}^{\mathrm{dev}}]\in\mathbb{R}^{2d_p}$. Both the projection matrix and reference mean are fixed during training. Further details are provided in Appendix~\ref{app:method-dual-jl}.

\subsection{Attribution-Oriented Step Representation Learning}

\textbf{View augmentation.} Before encoding, we independently augment each projected trajectory $\mathbf{X}_i={\bigl(\mathbf{x}_{i,t}\bigr)}_{t=1}^{T_i}$ twice to obtain two aligned views $\widetilde{\mathbf{X}}_i^{\,v}$, where $v\in\{\mathrm{f}_1,\mathrm{f}_2\}$. The input and augmented views share the same dimensions: $\mathbf{X}_i,\widetilde{\mathbf{X}}_i^{\,v}\in\mathbb{R}^{T_i\times 2d_p}$. For timestamp $t$ and feature coordinate $d\in\{1,\ldots,2d_p\}$, the augmented feature is
\begin{equation}
\label{eq:view-input}
\widetilde{x}_{i,t,d}^{\,v}
=
k_{i,t}^{v}\phi_{i,t,d}^{v}x_{i,t,d}
+
\epsilon_{i,t,d}^{v},
\qquad k_{i,t}^{v},\phi_{i,t,d}^{v}\in\{0,1\},
\qquad \epsilon_{i,t,d}^{v}\in\mathbb{R},
\end{equation}
where $k_{i,t}^{v}$ and $\phi_{i,t,d}^{v}$ are the step and feature masks, respectively, and $\epsilon_{i,t,d}^{v}$ is additive Gaussian noise. These perturbations are sampled independently for the two views. We protect the root-cause step at $t_i^*$ from step masking while allowing feature-level perturbations. The augmentation settings are reported in Appendix~\ref{app:hyperparameter-settings}.

\textbf{Encoding.} A bidirectional encoder $f_{\theta}$ maps each augmented view to $d_z$-dimensional step representations using the complete trajectory:
\begin{equation}
\label{eq:encoder-output}
\mathbf{Z}_i^{v}=f_{\theta}(\widetilde{\mathbf{X}}_i^{\,v})=\left(\mathbf{z}_{i,t}^{v}\right)_{t=1}^{T_i}\in\mathbb{R}^{T_i\times d_z}.
\end{equation}
A scoring head $g_{\psi}:\mathbb{R}^{d_z}\to\mathbb{R}$ produces the attribution score $\alpha_{i,t}^{v}=g_{\psi}(\mathbf{z}_{i,t}^{v})$ from each step representation $\mathbf{z}_{i,t}^{v}\in\mathbb{R}^{d_z}$, providing step-level supervision during encoder training.

\textbf{Root-cause contrastive objective.} For each root-cause step, we use its representation in view $\mathrm{f}_1$ as the anchor and the corresponding representation in view $\mathrm{f}_2$ as the sole positive, since both represent the same underlying step under different perturbations. All non-root steps in view $\mathrm{f}_2$ across the batch serve as negatives, promoting separation from diverse non-root contexts. The positive and negative cosine similarities are
\begin{equation}
\label{eq:root-cause-similarities}
\rho_{i,t_i^*}^{+}=\cos(\mathbf{z}_{i,t_i^*}^{\mathrm{f}_1},\mathbf{z}_{i,t_i^*}^{\mathrm{f}_2}),\qquad \rho_{i,t_i^*,j,t}^{-}=\cos(\mathbf{z}_{i,t_i^*}^{\mathrm{f}_1},\mathbf{z}_{j,t}^{\mathrm{f}_2}).
\end{equation}
Let $B_{\mathrm{fail}}=\{i\in B:\tau_i\in\mathcal{T}^{\mathrm{fail}}\}$ index the failed trajectories in the batch. We optimize these pairs with an InfoNCE objective:
\begin{equation}
\label{eq:root-cause-nce}
\mathcal{L}_{\mathrm{root}}=-\frac{1}{|B_{\mathrm{fail}}|}\sum_{i\in B_{\mathrm{fail}}}\log\frac{\exp(\rho_{i,t_i^*}^{+}/\kappa)}{\exp(\rho_{i,t_i^*}^{+}/\kappa)+\sum_{j\in B}\sum_{t\in\mathcal{N}(\tau_j)}\exp(\rho_{i,t_i^*,j,t}^{-}/\kappa)}.
\end{equation}
Here $\kappa>0$ is the temperature parameter. For successful trajectories, $\mathcal{N}(\tau_i)=\{1,\ldots,T_i\}$, so all their candidate steps contribute to the negative pool.

\textbf{Non-root-cause contrastive objective.} To learn discriminative non-root representations, we select anchors from the steps preceding the root cause. For each failed trajectory, let $\mathcal{P}_i=\{t\in\mathcal{N}(\tau_i):t<t_i^*\}$. When $\mathcal{P}_i$ is nonempty, the anchor timestamps are selected as
\begin{equation}
\label{eq:nonroot-anchor-selection}
\bar{\alpha}_{i,t}=\frac{\alpha_{i,t}^{\mathrm{f}_1}+\alpha_{i,t}^{\mathrm{f}_2}}{2},
\qquad
\mathcal{S}_i=\left\{\max\mathcal{P}_i,\ \operatorname*{arg\,max}_{t\in\mathcal{P}_i}\bar{\alpha}_{i,t}\right\}.
\end{equation}
The latest step provides a nearby reference, while the highest-scoring step is most likely to be mistaken for the root cause. We retain one anchor when selections coincide and choose the earliest step when scores tie. For each $t\in\mathcal{S}_i$, view $\mathrm{f}_1$ supplies the anchor, while view $\mathrm{f}_2$ supplies the same step as the positive and that trajectory's root-cause step as the sole negative:
\begin{equation}
\label{eq:nonroot-cause-similarities}
\rho_{i,t}^{+}=\cos(\mathbf{z}_{i,t}^{\mathrm{f}_1},\mathbf{z}_{i,t}^{\mathrm{f}_2}),\qquad \rho_{i,t}^{-}=\cos(\mathbf{z}_{i,t}^{\mathrm{f}_1},\mathbf{z}_{i,t_i^*}^{\mathrm{f}_2}).
\end{equation}
Let $B_{\mathrm{pre}}=\{i\in B_{\mathrm{fail}}:\mathcal{P}_i\neq\emptyset\}$ index the eligible failed trajectories. We average the InfoNCE losses first over the selected anchors within each trajectory and then over these trajectories:
\begin{equation}
\label{eq:nonroot-cause-nce}
\mathcal{L}_{\mathrm{non\text{-}root}}=-\frac{1}{|B_{\mathrm{pre}}|}\sum_{i\in B_{\mathrm{pre}}}\frac{1}{|\mathcal{S}_i|}\sum_{t\in\mathcal{S}_i}\log\frac{\exp(\rho_{i,t}^{+}/\kappa)}{\exp(\rho_{i,t}^{+}/\kappa)+\exp(\rho_{i,t}^{-}/\kappa)}.
\end{equation}
This reduction gives each eligible trajectory equal weight, regardless of whether it contributes one or two anchors. If $B_{\mathrm{pre}}$ is empty, the loss is zero.

\textbf{Listwise responsibility.} For each $i\in B_{\mathrm{fail}}$ and view $v$, we normalize step scores and maximize the root-cause probability at $t_i^*$. The ranking loss averages the two view-specific losses:
\begin{equation}
\label{eq:listwise-responsibility}
\mathcal{L}_{\mathrm{rank}}^{v}=-\frac{1}{|B_{\mathrm{fail}}|}\sum_{i\in B_{\mathrm{fail}}}\log\frac{\exp(\alpha_{i,t_i^*}^{v})}{\sum_{t=1}^{T_i}\exp(\alpha_{i,t}^{v})},
\qquad
\mathcal{L}_{\mathrm{rank}}=\frac{1}{2}\left(\mathcal{L}_{\mathrm{rank}}^{\mathrm{f}_1}+\mathcal{L}_{\mathrm{rank}}^{\mathrm{f}_2}\right).
\end{equation}
    
\textbf{Training objective.} The final encoder objective combines the trajectory-level ranking loss, the root-cause contrastive objective, and the non-root-cause contrastive objective:
\begin{equation}
\label{eq:training-objective}
\mathcal{L}=\lambda_{\mathrm{rank}}\mathcal{L}_{\mathrm{rank}}+\lambda_{\mathrm{root}}\mathcal{L}_{\mathrm{root}}+\lambda_{\mathrm{non\text{-}root}}\mathcal{L}_{\mathrm{non\text{-}root}},
\end{equation}
where $\lambda_{\mathrm{rank}}$, $\lambda_{\mathrm{root}}$, and $\lambda_{\mathrm{non\text{-}root}}$ are the loss weights. At inference, we encode the unaugmented trajectory and use the scoring head to identify the highest-scoring step as the predicted root cause. The specific optimization settings are reported in Appendix~\ref{app:hyperparameter-settings}.

% !TEX root = ../main.tex
\section{Experiments}

\subsection{Experimental Setup}
\label{sec:experimental-setup}

\textbf{Benchmarks.} We evaluate our method on four failure-attribution benchmarks: Who\&When~\citep{zhang2025whowhen}, Who\&When Pro~\citep{liu2026whowhenpro}, AFTraj-2K~\citep{zhang2026agentforesight}, and TraceElephant~\citep{chen2026traceElephant}. Our evaluation includes 124 algorithm-generated (AG) and 58 handcrafted (HC) trajectories from Who\&When, 1,251 test trajectories from an 8:2 task-grouped split of the Who\&When Pro text subset, 163 failed test trajectories from AFTraj-2K, and 175 valid trajectories from the Captain-Agent (85) and Magentic-One (90) subsets of TraceElephant. All TraceElephant results use these two subsets. Further benchmark details are provided in Appendix~\ref{app:benchmark-details}.

\textbf{Training datasets.} We use different training datasets for different target benchmarks. Specifically, for Who\&When and TraceElephant, we train on our ReCast-2K dataset. For AFTraj-2K, we train on its official training split. For Who\&When Pro, we use the training partition described above. See Appendix~\ref{app:dataset-construction} for construction and annotation details.

\textbf{Evaluation protocol.} We report \emph{Hit@$k$} for $k\in\{1,3,5\}$, \emph{Mean Reciprocal Rank (MRR)}, \emph{AUROC}, and \emph{AUPRC}. Hit@$k$ measures whether the annotated root-cause step appears among the top-$k$ predictions, while MRR captures its overall ranking position. AUROC and AUPRC assess score-level separation between root-cause and non-root steps, with AUPRC being particularly informative under class imbalance. All methods use the same three seeds, and we report their mean and standard deviation. See Appendix~\ref{app:evaluation-metrics} for detailed definitions.

\textbf{Baselines.} We group the baselines into three categories. \textbf{(1) Heuristic baselines.} First selects the first candidate step as the predicted root cause, whereas Random uniformly samples one candidate step. \textbf{(2) LLM-based methods.} We evaluate the all-at-once and step-by-step protocols from Who\&When~\citep{zhang2025whowhen} using Qwen3.5-27B~\citep{qwen3.5}, GPT-5.6 Luna~\citep{openai2026gpt56}, and DeepSeek V4 Flash~\citep{xu2026deepseek}, including both open and closed models. We also evaluate CHIEF~\citep{wang2026chief} with DeepSeek V4 Flash and include the reported results of AgenTracer~\citep{zhang2026agentracer}. \textbf{(3) Feature-based methods.} We compare with OAT~\citep{yeh2026oat}, StepFinder~\citep{zhu2026stepfinder}, MASPrism~\citep{liu2026masprism}, and ASCon~\citep{jiang2026ascon}. In each in-domain setting, the supervised feature-based baselines use the same failed-trajectory training split as our method. OAT instead follows its original positive-only training protocol. Detailed baseline settings are provided in Appendix~\ref{app:baseline-settings}.

\textbf{Implementation details.} We extract step representations using Qwen3.5-27B and set the projection dimension of each branch to $d_p=2{,}048$. The bidirectional Transformer encoder uses four layers with $d_z=128$. Training uses AdamW for up to 50 epochs with a batch size of 32 and a learning rate of $2\times10^{-4}$. The InfoNCE temperature is $\kappa=0.12$, and the loss weights are $\lambda_{\mathrm{rank}}=1.0$, $\lambda_{\mathrm{root}}=0.95$, and $\lambda_{\mathrm{non\text{-}root}}=0.25$. Additional settings are provided in Appendix~\ref{app:hyperparameter-settings}.

\begin{table}[t]
\centering
\small
\caption{Hit@1 attribution accuracy (\%) on Who\&When and TraceElephant. Results from three random seeds are summarized as mean $\pm$ population standard deviation. The LLMs are Qwen3.5-27B (Qwen), DeepSeek V4 Flash (DS), and GPT-5.6 Luna (GPT). Best results are shown in \textbf{bold}, and second-best results are \underline{underlined}.}
\label{tab:main-hc-ag}
\setlength{\tabcolsep}{0.8pt}
\renewcommand{\arraystretch}{1.25}
\begin{tabular*}{\textwidth}{@{\extracolsep{\fill}}llrrrr@{}}
\hline
\multirow{2}{*}{\textbf{Category}} & \multirow{2}{*}{\textbf{Method}} & \multicolumn{2}{c}{\textbf{Who\&When}} & \multicolumn{2}{c}{\textbf{TraceElephant}} \\
\cline{3-4}\cline{5-6}
& & \textbf{Algorithm} & \textbf{Handcrafted} & \textbf{Captain} & \textbf{Magentic} \\
\hline
\multirow{2}{*}{Heuristic} & First & 16.13$\pm$0.00 & 1.72$\pm$0.00 & 7.06$\pm$0.00 & 6.67$\pm$0.00 \\
& Random & 16.40$\pm$4.85 & 5.75$\pm$0.81 & 12.55$\pm$6.10 & 5.93$\pm$2.62 \\
\hline
\multirow{8}{*}{LLM-based} & All-at-Once (Qwen) & 26.61$\pm$0.00 & 3.45$\pm$0.00 & 7.06$\pm$0.00 & 4.44$\pm$0.00 \\
& All-at-Once (DS) & 22.04$\pm$0.76 & 10.34$\pm$2.44 & 12.55$\pm$1.11 & 9.63$\pm$0.52 \\
& All-at-Once (GPT) & 27.42$\pm$0.66 & 18.39$\pm$5.69 & 14.90$\pm$4.74 & 8.15$\pm$0.52 \\
& Step-by-Step (Qwen) & 32.26$\pm$0.00 & 17.24$\pm$0.00 & 25.88$\pm$0.00 & 18.89$\pm$0.00 \\
& Step-by-Step (DS) & 29.84$\pm$0.66 & 21.26$\pm$2.15 & 23.92$\pm$2.93 & 11.11$\pm$2.40 \\
& Step-by-Step (GPT) & 31.99$\pm$0.76 & 18.97$\pm$3.72 & 29.41$\pm$2.54 & 15.93$\pm$0.52 \\
& CHIEF (DS) & \underline{39.78$\pm$3.25} & \underline{24.14$\pm$3.72} & 23.53$\pm$2.54 & 25.19$\pm$1.89 \\
& AgenTracer$^\dagger$ & 37.30 & 20.68 & -- & -- \\
\hline
\multirow{4}{*}{Feature-based} & OAT & 9.68$\pm$0.66 & 5.17$\pm$5.08 & 3.53$\pm$0.00 & 17.41$\pm$1.39 \\
& MASPrism & 33.87$\pm$0.00 & 5.17$\pm$0.00 & 18.82$\pm$0.00 & 7.78$\pm$0.00 \\
& StepFinder & 22.85$\pm$2.12 & 18.97$\pm$2.44 & 27.84$\pm$2.00 & 19.26$\pm$0.52 \\
& ASCon & 26.34$\pm$1.01 & 21.84$\pm$0.81 & \underline{30.98$\pm$1.11} & \underline{25.93$\pm$0.52} \\
\hline
\textbf{Ours} & \textbf{ReCast} & \textbf{45.43}$\pm$\textbf{2.01} & \textbf{33.33}$\pm${\textbf{1.63}} & \textbf{35.69}$\pm$\textbf{2.93} & \textbf{27.04}$\pm$\textbf{1.89} \\
\hline
\end{tabular*}
\vspace{2pt}
\parbox{\textwidth}{\scriptsize $^\dagger$ \textit{AgenTracer-8B} is not publicly available; therefore, we cannot report multi-seed results.}
\end{table}

\subsection{Main Results}
\label{sec:main-results}

\textbf{Comparison on Who\&When and TraceElephant.} As shown in Table~\ref{tab:main-hc-ag}, our method achieves the best Hit@1 accuracy across all four subsets of Who\&When and TraceElephant. On Who\&When, it reaches $45.43\%$ on Algorithm and $33.33\%$ on Handcrafted, exceeding the second-best method, CHIEF, by 5.65 and 9.19 percentage points, respectively. On TraceElephant, our method reaches $35.69\%$ on Captain-Agent and $27.04\%$ on Magentic-One, outperforming the second-best method, ASCon, by 4.71 and 1.11 percentage points, respectively. These improvements support the effectiveness of learning attribution-oriented step representations for failure attribution.

\begin{table}[t]
\centering
\small
\caption{Hit@1 attribution accuracy (\%) on the two in-domain benchmark settings. Results from three random seeds are summarized as mean $\pm$ population standard deviation. Best results are shown in \textbf{bold}, and second-best results are \underline{underlined}.}
\label{tab:main-in-domain}
\setlength{\tabcolsep}{4pt}
\renewcommand{\arraystretch}{1.2}
\begin{tabular*}{\textwidth}{@{\extracolsep{\fill}}llrr@{}}
\hline
\textbf{Category} & \textbf{Method} & \textbf{AFTraj-2K} & \textbf{Who\&When Pro} \\
\hline
\multirow{2}{*}{Heuristic}
& First & 0.00$\pm$0.00 & 7.35$\pm$0.00 \\
& Random & 11.45$\pm$0.77 & 25.69$\pm$0.69 \\
\hline
\multirow{6}{*}{LLM-based}
& All-at-Once (Qwen) & 20.86$\pm$0.00 & 21.10$\pm$0.00 \\
& All-at-Once (DS) & 30.47$\pm$5.02 & 25.95$\pm$1.29 \\
& All-at-Once (GPT) & 19.02$\pm$0.87 & 33.33$\pm$1.27 \\
& Step-by-Step (Qwen) & 51.53$\pm$0.00 & 60.33$\pm$0.04 \\
& Step-by-Step (DS) & 53.99$\pm$1.74 & 62.16$\pm$0.26 \\
& Step-by-Step (GPT) & 54.40$\pm$0.77 & 63.58$\pm$0.37 \\
\hline
\multirow{4}{*}{Feature-based}
& OAT & 31.49$\pm$1.04 & 16.47$\pm$0.99 \\
& MASPrism & 22.09$\pm$0.00 & 10.87$\pm$0.00 \\
& StepFinder & 73.82$\pm$1.61 & 69.81$\pm$0.61 \\
& ASCon & \underline{83.03$\pm$0.29} & \underline{82.71$\pm$0.21} \\
\hline
\textbf{Ours} & \textbf{ReCast} & \textbf{84.05$\pm$0.87} & \textbf{91.42$\pm$0.21} \\
\hline
\end{tabular*}
\end{table}

\textbf{Comparison on AFTraj-2K and Who\&When Pro.} Table~\ref{tab:main-in-domain} reports the Hit@1 results on AFTraj-2K and Who\&When Pro. On AFTraj-2K, our method achieves $84.05\%$ Hit@1, outperforming GPT-5.6 Luna with Step-by-Step and ASCon by 29.65 and 1.02 percentage points, respectively. On Who\&When Pro, it reaches $91.42\%$, exceeding the same baselines by 27.84 and 8.71 points. These gains show that our method remains effective when trained on other datasets, suggesting that its benefits are not limited to ReCast-2K.

\begin{figure}[t]
\centering
\includegraphics[width=\textwidth]{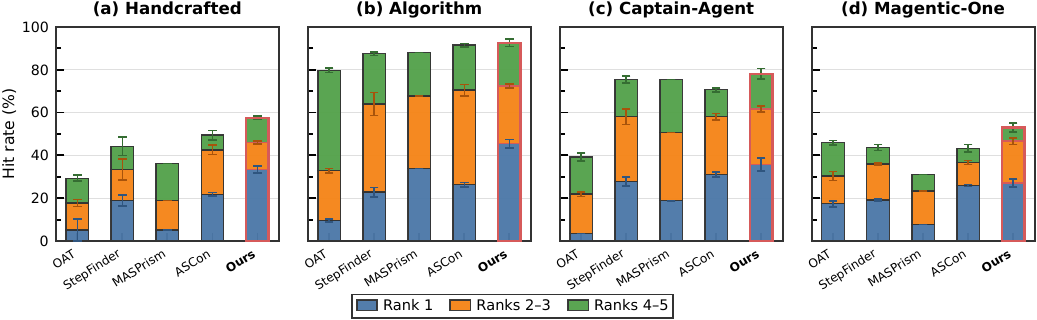}
\caption{Cumulative Hit@$k$ accuracy on Who\&When and TraceElephant. Stacked segments denote ranks 1, 2--3, and 4--5; error bars show the population standard deviation over three seeds.}
\label{fig:whowhen-traceelephant-hitk}
\end{figure}

\textbf{Hit@$k$ comparison.} Figure~\ref{fig:whowhen-traceelephant-hitk} compares cumulative Hit@1, Hit@3, and Hit@5 across methods on Who\&When and TraceElephant. We make two observations: \textbf{(1) Strong top-$k$ coverage.} Our method achieves the highest Hit@5 accuracy across all four benchmark settings, maintaining its advantage under a more relaxed ranking criterion. \textbf{(2) Better top-rank prioritization.} On Captain-Agent, our method reaches $78.04\%$ Hit@5, exceeding MASPrism's $75.29\%$, while achieving a substantially higher Hit@1 of $35.69\%$ compared with $18.82\%$ for MASPrism. These results indicate that our method ranks the root-cause step more effectively, placing it first more often while maintaining strong top-five coverage.

\textbf{Learned representation separation.} To assess whether our method addresses the limited separation identified in Section~\ref{sec:observation}, we evaluate the learned step representations using the same metrics. Compared with the basic representations, the learned representations achieve lower ARS and GRS and higher SI on average across seeds. On Handcrafted and Algorithm, mean SI improves from the best basic-representation values of $0.17$ and $0.03$ to $0.59$ and $0.29$, respectively. These results indicate improved within-trace and cross-trace separation, supporting our goal of learning attribution-oriented step representations. See Appendix~\ref{app:trained-representation-separation} for the full comparison.

% !TEX root = ../main.tex
\section{Analysis}

\label{sec:ablation-study}

\begin{table}[t]
\centering
\small
\caption{Ablation results on Who\&When. Hit@1 (\%) over all trajectories is reported as mean $\pm$ standard deviation across three seeds. $\Delta$ Hit@1 reports the change from the full-model mean in percentage points (pp).}
\label{tab:ablation-study}
\setlength{\tabcolsep}{5pt}
\renewcommand{\arraystretch}{1.02}
\begin{tabular*}{\textwidth}{@{\extracolsep{\fill}}llcr@{}}
\toprule
\textbf{Component} & \textbf{Variant} & \textbf{Hit@1} & \textbf{$\Delta$ Hit@1 (pp)} \\
\midrule
None & \textbf{Full model} & \textbf{41.58}$\pm$\textbf{1.04} & --- \\
\midrule
\multirow{5}{*}{Layer selection}
& All 64 layers & 38.83$\pm$2.30 & $-2.75$ \\
& Uniform 8 & 38.10$\pm$2.55 & $-3.48$ \\
& Last 8 & 37.55$\pm$0.93 & $-4.03$ \\
& Random 8 & 37.18$\pm$0.69 & $-4.40$ \\
& Global probe top-8 & 38.83$\pm$1.13 & $-2.75$ \\
\midrule
\multirow{2}{*}{Projection branches}
& Pattern only & 40.11$\pm$1.96 & $-1.47$ \\
& Deviation only & 41.03$\pm$2.92 & $-0.55$ \\
\midrule
\multirow{2}{*}{Encoder configuration}
& Per-step MLP & 40.48$\pm$2.12 & $-1.10$ \\
& No augmentation & 37.36$\pm$0.90 & $-4.22$ \\
\midrule
\multirow{2}{*}{Non-root anchor selection}
& Latest only & 39.93$\pm$0.93 & $-1.65$ \\
& Highest-score only & 39.74$\pm$0.52 & $-1.84$ \\
\midrule
\multirow{3}{*}{Objective}
& Rank only & 37.73$\pm$0.26 & $-3.85$ \\
& Rank + root & 40.48$\pm$0.69 & $-1.10$ \\
& Rank + non-root & 37.73$\pm$1.04 & $-3.85$ \\
\bottomrule
\end{tabular*}
\end{table}

\textbf{Ablation Study.} As summarized in Table~\ref{tab:ablation-study}, we conduct ablation experiments on the Who\&When benchmark. Each ablation varies one component while keeping the remaining settings fixed. The full model achieves the highest mean Hit@1 of $41.58\%$. Attribution-aware layer selection outperforms five alternatives by 2.75--4.40 percentage points, while the joint loss improves over ranking-only training by 3.85 points. Removing augmentation reduces Hit@1 by 4.22 points, highlighting its contribution to representation learning. Ablations of the projection branches, encoder, and non-root anchor selection also yield lower accuracy, supporting the overall design.

\textbf{Backbone and efficiency analyses.} In Appendix~\ref{app:additional-backbone-comparison}, we compare Qwen3.5 backbones with 0.8B, 9B, and 27B parameters while keeping the downstream architecture fixed. The 27B model achieves the highest Hit@1 on Who\&When, AFTraj-2K, and Who\&When Pro, whereas 0.8B and 9B perform best on Captain-Agent and Magentic-One, respectively. These results indicate that larger backbones can improve attribution accuracy, but increasing model size does not consistently yield better performance. Appendix~\ref{app:time-efficiency} further compares accuracy and processing time, demonstrating efficient attribution with ReCast when step features are precomputed.

% !TEX root = ../main.tex
\section{Conclusion}

This work explores step representation learning for failure attribution in LLM-based agent systems. To understand how well existing representations distinguish root-cause steps from other steps, we conduct an empirical study of text embeddings and hidden states, finding limited separation under the evaluated metrics. Motivated by this observation, ReCast transforms frozen LLM hidden states into attribution-oriented step representations through layer selection, complementary pattern and deviation feature construction, and contextual representation learning with contrastive and ranking objectives. We introduce ReCast-2K to support training. ReCast achieves the best Hit@1 across four benchmarks, while representation analysis shows improved within-trace and cross-trace separation. These results highlight the value of learning step representations tailored to failure attribution.

% !TEX root = ../main.tex
\subsection*{AI use statement}

In this work, we used generative AI tools for dataset construction and method implementation.
We have not used generative AI tools for writing mathematical proofs,
and tasks involving critical ingredients for proofs are not applicable to this work.
Additionally, we used generative AI tools for drafting sections of the manuscript, language editing, and reference formatting. We have reviewed all AI-assisted work. We checked the accuracy and consistency of AI-assisted content against experimental records and the implementation. We take responsibility for the final content of this work, including text, claims or artifacts produced with the aid of generative AI.

\subsection*{Reproducibility statement}

We provide our code to support reproducibility. Appendices~\ref{app:benchmark-details} and~\ref{app:dataset-construction} describe the datasets, candidate-step processing, and training-data construction. Method implementation and evaluation settings, including metrics, hyperparameters, and random seeds, are detailed in Appendices~\ref{app:method-details} and~\ref{app:evaluation-details}. Baseline configurations and LLM attribution prompts appear in Appendices~\ref{app:baseline-settings} and~\ref{app:baseline-prompts}, respectively.

\bibliography{reference}
\bibliographystyle{iclr2027_conference}

\clearpage
\appendix
% !TEX root = ../main.tex

\noindent{\Large\scshape Appendix\par}
\vspace{1.5ex}

\section*{Contents}

\begingroup
\setlength{\parindent}{0pt}
\hyperref[app:empirical-similarity-study]{\textbf{\ref*{app:empirical-similarity-study}\quad Empirical Similarity Study}}\dotfill\hyperref[app:empirical-similarity-study]{\pageref*{app:empirical-similarity-study}}\par
\hspace{1.5em}\hyperref[app:empirical-similarity-metrics]{\ref*{app:empirical-similarity-metrics}\quad Evaluation Metrics}\dotfill\hyperref[app:empirical-similarity-metrics]{\pageref*{app:empirical-similarity-metrics}}\par
\hspace{1.5em}\hyperref[app:empirical-similarity-representations]{\ref*{app:empirical-similarity-representations}\quad Representations}\dotfill\hyperref[app:empirical-similarity-representations]{\pageref*{app:empirical-similarity-representations}}\par
\hspace{1.5em}\hyperref[app:empirical-similarity-results]{\ref*{app:empirical-similarity-results}\quad Complete Results}\dotfill\hyperref[app:empirical-similarity-results]{\pageref*{app:empirical-similarity-results}}\par
\hyperref[app:method-details]{\textbf{\ref*{app:method-details}\quad Method Implementation Details}}\dotfill\hyperref[app:method-details]{\pageref*{app:method-details}}\par
\hspace{1.5em}\hyperref[app:method-layer-selection]{\ref*{app:method-layer-selection}\quad Attribution-Aware Layer Selection}\dotfill\hyperref[app:method-layer-selection]{\pageref*{app:method-layer-selection}}\par
\hspace{1.5em}\hyperref[app:method-dual-jl]{\ref*{app:method-dual-jl}\quad Dual-branch Projection}\dotfill\hyperref[app:method-dual-jl]{\pageref*{app:method-dual-jl}}\par
\hyperref[app:benchmark-details]{\textbf{\ref*{app:benchmark-details}\quad Benchmark Details}}\dotfill\hyperref[app:benchmark-details]{\pageref*{app:benchmark-details}}\par
\hspace{1.5em}\hyperref[app:benchmark-whowhen]{\ref*{app:benchmark-whowhen}\quad Who\&When}\dotfill\hyperref[app:benchmark-whowhen]{\pageref*{app:benchmark-whowhen}}\par
\hspace{1.5em}\hyperref[app:benchmark-whowhen-pro]{\ref*{app:benchmark-whowhen-pro}\quad Who\&When Pro}\dotfill\hyperref[app:benchmark-whowhen-pro]{\pageref*{app:benchmark-whowhen-pro}}\par
\hspace{1.5em}\hyperref[app:benchmark-aftraj]{\ref*{app:benchmark-aftraj}\quad AFTraj-2K}\dotfill\hyperref[app:benchmark-aftraj]{\pageref*{app:benchmark-aftraj}}\par
\hspace{1.5em}\hyperref[app:benchmark-traceelephant]{\ref*{app:benchmark-traceelephant}\quad TraceElephant}\dotfill\hyperref[app:benchmark-traceelephant]{\pageref*{app:benchmark-traceelephant}}\par
\hspace{1.5em}\hyperref[app:benchmark-data-processing]{\ref*{app:benchmark-data-processing}\quad Data Processing}\dotfill\hyperref[app:benchmark-data-processing]{\pageref*{app:benchmark-data-processing}}\par
\vspace{0.25em}
\hyperref[app:dataset-construction]{\textbf{\ref*{app:dataset-construction}\quad Training Dataset Construction}}\dotfill\hyperref[app:dataset-construction]{\pageref*{app:dataset-construction}}\par
\hspace{1.5em}\hyperref[app:dataset-task-generation]{\ref*{app:dataset-task-generation}\quad Task Sampling and Trajectory Generation}\dotfill\hyperref[app:dataset-task-generation]{\pageref*{app:dataset-task-generation}}\par
\hspace{1.5em}\hyperref[app:dataset-root-cause-annotation]{\ref*{app:dataset-root-cause-annotation}\quad Root-Cause Annotation}\dotfill\hyperref[app:dataset-root-cause-annotation]{\pageref*{app:dataset-root-cause-annotation}}\par
\hspace{1.5em}\hyperref[app:dataset-statistics]{\ref*{app:dataset-statistics}\quad Final Dataset Statistics}\dotfill\hyperref[app:dataset-statistics]{\pageref*{app:dataset-statistics}}\par
\hyperref[app:evaluation-details]{\textbf{\ref*{app:evaluation-details}\quad Evaluation Details and Hyperparameters}}\dotfill\hyperref[app:evaluation-details]{\pageref*{app:evaluation-details}}\par
\hspace{1.5em}\hyperref[app:evaluation-metrics]{\ref*{app:evaluation-metrics}\quad Evaluation Metrics}\dotfill\hyperref[app:evaluation-metrics]{\pageref*{app:evaluation-metrics}}\par
\hspace{1.5em}\hyperref[app:hyperparameter-settings]{\ref*{app:hyperparameter-settings}\quad Hyperparameter Settings}\dotfill\hyperref[app:hyperparameter-settings]{\pageref*{app:hyperparameter-settings}}\par
\hyperref[app:baseline-settings]{\textbf{\ref*{app:baseline-settings}\quad Baseline Settings}}\dotfill\hyperref[app:baseline-settings]{\pageref*{app:baseline-settings}}\par
\hspace{1.5em}\hyperref[app:baseline-basic]{\ref*{app:baseline-basic}\quad Heuristic Baselines}\dotfill\hyperref[app:baseline-basic]{\pageref*{app:baseline-basic}}\par
\hspace{1.5em}\hyperref[app:baseline-llm]{\ref*{app:baseline-llm}\quad LLM-Based Methods}\dotfill\hyperref[app:baseline-llm]{\pageref*{app:baseline-llm}}\par
\hspace{1.5em}\hyperref[app:baseline-small-models]{\ref*{app:baseline-small-models}\quad Feature-Based Methods}\dotfill\hyperref[app:baseline-small-models]{\pageref*{app:baseline-small-models}}\par
\hyperref[app:additional-experiments]{\textbf{\ref*{app:additional-experiments}\quad Additional Experimental Results}}\dotfill\hyperref[app:additional-experiments]{\pageref*{app:additional-experiments}}\par
\hspace{1.5em}\hyperref[app:additional-mrr]{\ref*{app:additional-mrr}\quad MRR Across Rank Cutoffs}\dotfill\hyperref[app:additional-mrr]{\pageref*{app:additional-mrr}}\par
\hspace{1.5em}\hyperref[app:additional-score-discrimination]{\ref*{app:additional-score-discrimination}\quad Score-Level Discrimination}\dotfill\hyperref[app:additional-score-discrimination]{\pageref*{app:additional-score-discrimination}}\par
\hspace{1.5em}\hyperref[app:additional-backbone-comparison]{\ref*{app:additional-backbone-comparison}\quad Hidden-State Backbone Comparison}\dotfill\hyperref[app:additional-backbone-comparison]{\pageref*{app:additional-backbone-comparison}}\par
\hspace{1.5em}\hyperref[app:time-efficiency]{\ref*{app:time-efficiency}\quad Time Efficiency Analysis}\dotfill\hyperref[app:time-efficiency]{\pageref*{app:time-efficiency}}\par
\hyperref[app:case-study]{\textbf{\ref*{app:case-study}\quad Case Study}}\dotfill\hyperref[app:case-study]{\pageref*{app:case-study}}\par
\hyperref[app:baseline-prompts]{\textbf{\ref*{app:baseline-prompts}\quad Prompt Templates}}\dotfill\hyperref[app:baseline-prompts]{\pageref*{app:baseline-prompts}}\par
\hspace{1.5em}\hyperref[app:all-at-once-prompt]{\ref*{app:all-at-once-prompt}\quad All-at-Once Prompt}\dotfill\hyperref[app:all-at-once-prompt]{\pageref*{app:all-at-once-prompt}}\par
\hspace{1.5em}\hyperref[app:step-by-step-prompt]{\ref*{app:step-by-step-prompt}\quad Step-by-Step Prompt}\dotfill\hyperref[app:step-by-step-prompt]{\pageref*{app:step-by-step-prompt}}\par
\endgroup

% !TEX root = ../../main.tex
\section{Empirical Similarity Study}
\label{app:empirical-similarity-study}

\subsection{Evaluation Metrics}
\label{app:empirical-similarity-metrics}

In this subsection, $\mathbf{x}_{i,t}$ denotes the representation being evaluated for step $t$ in failed trace $\tau_i$, including text embeddings, hidden states, projected features, or learned representations. We use the root-cause timestamps $t_i^*$ and non-root sets $\mathcal{N}(\tau_i)$ defined in Section~\ref{sec:problem-definition}. Let $\mathcal{T}_{\mathrm{pre}}=\{\tau_i\in\mathcal{T}^{\mathrm{fail}}:t_i^*>1\}$ contain the traces whose root cause has a preceding step. We define Adjacent Root Similarity (ARS) as
\begin{equation}
\label{eq:observation-ars}
\mathrm{ARS}
=
\frac{1}{|\mathcal{T}_{\mathrm{pre}}|}
\sum_{\tau_i\in\mathcal{T}_{\mathrm{pre}}}
\cos\!\left(
\mathbf{x}_{i,t_i^*},
\mathbf{x}_{i,t_i^*-1}
\right).
\end{equation}
Global Root Similarity (GRS) is defined as
\begin{equation}
\label{eq:observation-grs}
\mathrm{GRS}
=
\frac{1}{|\mathcal{T}^{\mathrm{fail}}|}
\sum_{\tau_i\in\mathcal{T}^{\mathrm{fail}}}
\frac{1}{|\mathcal{N}(\tau_i)|}
\sum_{t\in\mathcal{N}(\tau_i)}
\cos\!\left(
\mathbf{x}_{i,t_i^*},
\mathbf{x}_{i,t}
\right).
\end{equation}
For the cosine Silhouette Index (SI)~\citep{rousseeuw1987silhouettes}, we pool the root-cause representations into $\mathcal{X}_{\mathrm{root}}=\{\mathbf{x}_{i,t_i^*}:\tau_i\in\mathcal{T}^{\mathrm{fail}}\}$ and the non-root representations into $\mathcal{X}_{\mathrm{non}}=\{\mathbf{x}_{i,t}:\tau_i\in\mathcal{T}^{\mathrm{fail}},\,t\in\mathcal{N}(\tau_i)\}$. For each representation $\mathbf{x}$, we define its same-class and opposite-class sets as
\[
\bigl(\mathcal{S}(\mathbf{x}),\mathcal{O}(\mathbf{x})\bigr)
=
\begin{cases}
\bigl(\mathcal{X}_{\mathrm{root}},\mathcal{X}_{\mathrm{non}}\bigr), & \mathbf{x}\in\mathcal{X}_{\mathrm{root}},\\
\bigl(\mathcal{X}_{\mathrm{non}},\mathcal{X}_{\mathrm{root}}\bigr), & \mathbf{x}\in\mathcal{X}_{\mathrm{non}}.
\end{cases}
\]
Using cosine distance $d_{\cos}(\mathbf{x},\mathbf{z})=1-\cos(\mathbf{x},\mathbf{z})$, we compute
\begin{equation}
\label{eq:observation-si-distances}
\begin{aligned}
a(\mathbf{x})
&=
\frac{1}{|\mathcal{S}(\mathbf{x})|-1}
\sum_{\mathbf{z}\in\mathcal{S}(\mathbf{x})\setminus\{\mathbf{x}\}}
d_{\cos}\!\left(\mathbf{x},\mathbf{z}\right),\\
b(\mathbf{x})
&=
\frac{1}{|\mathcal{O}(\mathbf{x})|}
\sum_{\mathbf{z}\in\mathcal{O}(\mathbf{x})}
d_{\cos}\!\left(\mathbf{x},\mathbf{z}\right).
\end{aligned}
\end{equation}
The resulting SI is
\begin{equation}
\label{eq:observation-si}
\mathrm{SI}
=
\frac{1}{|\mathcal{X}_{\mathrm{root}}|+|\mathcal{X}_{\mathrm{non}}|}
\sum_{\mathbf{x}\in\mathcal{X}_{\mathrm{root}}\cup\mathcal{X}_{\mathrm{non}}}
\frac{b(\mathbf{x})-a(\mathbf{x})}
{\max\!\left(a(\mathbf{x}),b(\mathbf{x})\right)}.
\end{equation}
Together, these metrics assess two aspects of representation quality: within-trace and cross-trace separation. Within a trace, ARS measures similarity between the root-cause step and its immediately preceding step, while GRS measures its average similarity to all other steps; lower values indicate stronger separation. Across traces, SI measures the separation between root-cause and non-root representations, with higher values indicating better separation.

\subsection{Representations}
\label{app:empirical-similarity-representations}

We compare the following three representations of candidate model-action steps:
\begin{itemize}[leftmargin=1em]
\item \textbf{Text embedding.} Following StepFinder, we encode the serialized text of each candidate model-action step with Qwen3-Embedding-0.6B, producing a $1{,}024$-dimensional vector.
\item \textbf{Mean last-layer hidden state.} Following MASPrism~\citep{liu2026masprism} in using LLM internal signals, we average the Qwen3.5-27B last-layer hidden states over action tokens at each candidate timestamp, producing a $5{,}120$-dimensional vector.
\item \textbf{Last-token last-layer hidden state.} We use the Qwen3.5-27B last-layer hidden state of the final token at each candidate timestamp, producing a $5{,}120$-dimensional vector.
\end{itemize}

\subsection{Complete Results}
\label{app:empirical-similarity-results}
\label{app:trained-representation-separation}

\begin{table}[t]
\centering
\small
\caption{Separation of step representations measured by ARS, GRS, and SI. ReCast results show mean $\pm$ population standard deviation across seeds $6$, $20$, and $42$; other representations are fixed. Best results are bolded for each dataset and metric, based on mean values for ReCast.}
\label{tab:observation-results}
\renewcommand{\arraystretch}{1.12}
\setlength{\tabcolsep}{4pt}
\begin{tabular*}{\textwidth}{@{\extracolsep{\fill}}llccc@{}}
\toprule
Dataset & Representation & ARS $\downarrow$ & GRS $\downarrow$ & SI $\uparrow$ \\
\midrule
\multirow{5}{*}{Training} & Text embedding & $0.5478$ & $0.6173$ & $0.1482$ \\
& Hidden[-1] (Mean) & $0.7254$ & $0.7630$ & $0.0131$ \\
& Hidden[-1] (Last) & $0.7832$ & $0.7686$ & $0.0228$ \\
& Projected features & $0.5763$ & $0.6001$ & $0.0566$ \\
& \textbf{ReCast} & $\mathbf{-0.2590}\pm0.2116$ & $\mathbf{-0.2749}\pm0.1977$ & $\mathbf{0.7032}\pm0.1804$ \\
\midrule
\multirow{5}{*}{Handcrafted} & Text embedding & $0.5195$ & $0.5885$ & $0.0959$ \\
& Hidden[-1] (Mean) & $0.7710$ & $0.7691$ & $0.0464$ \\
& Hidden[-1] (Last) & $0.7019$ & $0.6760$ & $0.1735$ \\
& Projected features & $0.6001$ & $0.6127$ & $0.0593$ \\
& \textbf{ReCast} & $\mathbf{0.4157}\pm0.0781$ & $\mathbf{0.4235}\pm0.0933$ & $\mathbf{0.5913}\pm0.1233$ \\
\midrule
\multirow{5}{*}{Algorithm} & Text embedding & $0.8334$ & $0.7223$ & $0.0309$ \\
& Hidden[-1] (Mean) & $0.8969$ & $0.8417$ & $-0.0409$ \\
& Hidden[-1] (Last) & $0.8049$ & $0.7944$ & $-0.0137$ \\
& Projected features & $0.7658$ & $0.6888$ & $0.0005$ \\
& \textbf{ReCast} & $\mathbf{0.4087}\pm0.0342$ & $\mathbf{0.3878}\pm0.0467$ & $\mathbf{0.2873}\pm0.0986$ \\
\bottomrule
\end{tabular*}
\end{table}

We evaluate the three basic representations alongside our encoder inputs and learned step representations on the same failed trajectories using the metrics defined above. In Table~\ref{tab:observation-results}, \emph{Projected features} denotes $\mathbf{x}_{i,t}\in\mathbb{R}^{2d_p}$, the concatenated pattern and deviation features from Eq.~\ref{eq:dual-jl-output} before augmentation and encoding, with $2d_p=4{,}096$. \emph{ReCast} denotes the learned contextualized step representations before the scoring head, with $d_z=128$.

Among the basic representations, text embeddings perform best in most comparisons, although their advantage varies by dataset and metric. All three representations exhibit substantial within-trace similarity, reflected in high ARS and GRS, and limited cross-trace separation, reflected in low SI. This pattern is consistent with the limited separation observed under these metrics in Section~\ref{sec:observation}.

In contrast, ReCast achieves the lowest mean ARS and GRS and the highest mean SI on every dataset. Its mean SI reaches $0.5913$ on Handcrafted and $0.2873$ on Algorithm, exceeding the best basic-representation values of $0.1735$ and $0.0309$, respectively. The gains over projected features further show that representation learning improves separation beyond feature construction. These improvements within and across trajectories provide quantitative support for learning attribution-oriented step representations.

% !TEX root = ../../main.tex

\section{Method Implementation Details}
\label{app:method-details}

\subsection{Attribution-Aware Layer Selection}
\label{app:method-layer-selection}

Section~\ref{sec:input-representation} presents the overall attribution-aware layer-selection procedure. Here, we detail fold construction, probe statistics, and layer ranking based on held-out evaluation results. We partition the failed training trajectories $\mathcal{T}^{\mathrm{fail}}$ into five folds using a deterministic hash of each trace identifier. For fold $f$, let $\mathcal{I}^{f}$ denote the held-out trajectories and $\mathcal{I}^{-f}=\mathcal{T}^{\mathrm{fail}}\setminus\mathcal{I}^{f}$ the trajectories in the other four folds. The probe is fitted on $\mathcal{I}^{-f}$ and evaluated on $\mathcal{I}^{f}$. The mean and sample-variance vectors used in Eq.~\ref{eq:layer-probe-direction-main} are
\begin{equation}
\label{eq:layer-probe-statistics}
\boldsymbol{\mu}^{(\ell)}
=
\frac{1}{|\mathcal{I}^{-f}|}
\sum_{\tau_i\in\mathcal{I}^{-f}}\mathbf{d}_{i}^{(\ell)},
\qquad
\mathbf{v}^{(\ell)}
=
\frac{1}{|\mathcal{I}^{-f}|-1}
\sum_{\tau_i\in\mathcal{I}^{-f}}
\left(\mathbf{d}_{i}^{(\ell)}-\boldsymbol{\mu}^{(\ell)}\right)
\odot
\left(\mathbf{d}_{i}^{(\ell)}-\boldsymbol{\mu}^{(\ell)}\right).
\end{equation}
Using the score $r_{i,t}^{(\ell)}$ defined in the main text, the within-trajectory rank of the labeled root cause is
\begin{equation}
\label{eq:layer-probe-root-rank}
\operatorname{rank}_i^{(\ell)}
=
1+
\sum_{t\in\mathcal{N}(\tau_i)}
\mathbf{1}
\left[
 r_{i,t}^{(\ell)}>r_{i,t_i^*}^{(\ell)}
\right],
\qquad
\forall\,\tau_i\in\mathcal{I}^{f}.
\end{equation}
Thus, $\operatorname{rank}_i^{(\ell)}=1$ when no non-root step has a strictly larger score than the root cause, with tied scores leaving the rank unchanged. We pool the out-of-fold results to evaluate each layer by Hit@1, MRR, and mean root margin:
\begin{align}
\label{eq:layer-probe-hit-one}
\mathrm{Hit@1}^{(\ell)}
&=
\frac{1}{\sum_{f=1}^{5}|\mathcal{I}^{f}|}
\sum_{f=1}^{5}
\sum_{\tau_i\in\mathcal{I}^{f}}
\mathbf{1}\left[\operatorname{rank}_i^{(\ell)}=1\right],\\
\label{eq:layer-probe-mrr}
\mathrm{MRR}^{(\ell)}
&=
\frac{1}{\sum_{f=1}^{5}|\mathcal{I}^{f}|}
\sum_{f=1}^{5}
\sum_{\tau_i\in\mathcal{I}^{f}}
\frac{1}{\operatorname{rank}_i^{(\ell)}},\\
\label{eq:layer-probe-margin}
\mathrm{Margin}^{(\ell)}
&=
\frac{1}{\sum_{f=1}^{5}|\mathcal{I}^{f}|}
\sum_{f=1}^{5}
\sum_{\tau_i\in\mathcal{I}^{f}}
\left(
 r_{i,t_i^*}^{(\ell)}
 -
 \max_{t\in\mathcal{N}(\tau_i)}
 r_{i,t}^{(\ell)}
\right).
\end{align}
To ensure coverage across model depth, we partition the $L$ layers into $k$ contiguous groups:
\begin{equation}
\label{eq:layer-probe-groups}
\mathcal{G}_b
=
\left\{
\left\lfloor\frac{(b-1)L}{k}\right\rfloor+1,
\ldots,
\left\lfloor\frac{bL}{k}\right\rfloor
\right\},
\quad b=1,\ldots,k.
\end{equation}
Within each group, we rank layers by Hit@1, breaking ties by MRR and then mean root margin, and select the highest-ranked layer:
\begin{equation}
\label{eq:layer-probe-stratified-selection}
\ell_b
=
\operatorname*{arg\,max}_{\ell\in\mathcal{G}_b}
\left(
\mathrm{Hit@1}^{(\ell)},
\mathrm{MRR}^{(\ell)},
\mathrm{Margin}^{(\ell)}
\right).
\end{equation}
The resulting layers and probe settings are reported in Appendix~\ref{app:hyperparameter-settings}.

\subsection{Dual-branch Projection}
\label{app:method-dual-jl}

Equation~\ref{eq:dual-jl-output} defines the pattern and deviation branches. Here we provide the implementation details not specified in the main text: how the sparse projection is sampled and the exact normalization and reference statistics. We sample a sparse projection over all $L$ layers and retain the columns corresponding to the selected layers, ensuring that layer-selection variants share the same projection coordinates. The random entries follow
\begin{equation}
\label{eq:dual-jl-entry-distribution}
\left(\mathbf{W}_{\mathrm{JL}}\right)_{a,b}
\sim
\begin{cases}
\dfrac{1}{\sqrt{\delta d_p}},
& \text{with probability } \dfrac{\delta}{2},\\[5pt]
0,
& \text{with probability } 1-\delta,\\[5pt]
-\dfrac{1}{\sqrt{\delta d_p}},
& \text{with probability } \dfrac{\delta}{2}.
\end{cases}
\qquad
\delta=\frac{1}{\sqrt{Ld_h}}.
\end{equation}
Here, $\mathbf{W}_{\mathrm{JL}}\in\mathbb{R}^{d_p\times kd_h}$, where $L$ is the total number of backbone layers ($L=64$ for Qwen3.5-27B), $d_h$ is the feature dimension of one hidden layer, $d_p$ is the projection dimension, and $\delta$ is the projection density, i.e., the probability that an entry of $\mathbf{W}_{\mathrm{JL}}$ is nonzero. The full-layer matrix has shape $d_p\times Ld_h$ and is generated with random state 42. Its random entries are sampled independently of both the attribution labels and the hidden-state covariance, and the retained matrix is fixed and shared by both branches and across training seeds.

All instances of $\mathrm{LN}$ in Eq.~\ref{eq:dual-jl-output} use parameter-free normalization. For a vector $\mathbf{v}\in\mathbb{R}^{n}$, where $n=d_h$ for a layer feature and $n=d_p$ for a projected branch output, we compute
\begin{equation}
\label{eq:dual-jl-parameter-free-ln}
\mathrm{LN}(\mathbf{v})
=
\frac{\mathbf{v}-\mathbb{E}[\mathbf{v}]}
{\sqrt{\operatorname{Var}(\mathbf{v})+\epsilon}}.
\end{equation}
Here, the expectation and variance are computed independently for each step over its $n$ feature coordinates, and $\epsilon$ is a numerical-stability constant. The normalization has no learned scale or bias.

The successful-reference mean $\boldsymbol{\mu}^{\mathrm{succ}}\in\mathbb{R}^{kd_h}$ used in Eq.~\ref{eq:dual-jl-output} is computed over all candidate steps in the successful training trajectories:
\begin{equation}
\label{eq:dual-jl-success-mean}
\boldsymbol{\mu}^{\mathrm{succ}}
=
\frac{
 \sum_{\tau_i\in\mathcal{T}^{\mathrm{succ}}}
 \sum_{t=1}^{T_i}\mathbf{h}_{i,t}
}{
 \sum_{\tau_i\in\mathcal{T}^{\mathrm{succ}}}T_i
}.
\end{equation}
Using the fixed projection matrix $\mathbf{W}_{\mathrm{JL}}$ and successful-reference mean $\boldsymbol{\mu}^{\mathrm{succ}}$, Eq.~\ref{eq:dual-jl-output} produces the pattern and deviation features $\mathbf{x}_{i,t}^{\mathrm{pat}}$ and $\mathbf{x}_{i,t}^{\mathrm{dev}}$, which are concatenated directly without additional weighting or branch-specific scoring heads. The remaining dimensions and hyperparameters are provided in Appendix~\ref{app:hyperparameter-settings}.

% !TEX root = ../../main.tex
\section{Benchmark Details}
\label{app:benchmark-details}

\subsection{Who\&When}
\label{app:benchmark-whowhen}

Who\&When~\citep{zhang2025whowhen} contains 184 failed trajectories generated under two MAS construction settings and evaluated on two task benchmarks. The algorithm-generated (AG) setting uses CaptainAgent from the AG2 library~\citep{microsoft2024ag2} to construct a task-specific MAS for each query, whereas the handcrafted (HC) setting uses the fixed, manually designed Magentic-One system~\citep{fourney2024magentic}; both settings use GPT-4o~\citep{hurst2024gpt} as the base language model. The underlying tasks are drawn from GAIA~\citep{mialon2024gaia}, which evaluates agents on real-world questions requiring reasoning and tool use, and AssistantBench~\citep{yoran2024assistantbench}, which focuses on realistic and time-consuming web tasks.

The AG subset contains 126 failed trajectories, including 98 generated on GAIA and 28 on AssistantBench. The HC subset contains 58 failed trajectories produced by Magentic-One, including 30 trajectories from GAIA and 28 from AssistantBench. Each trajectory is annotated with the failure-responsible agent, the decisive error step, and a natural-language explanation. Because our candidate set is restricted to model-generated actions, we exclude two AG trajectories whose annotated decisive errors correspond to outputs from the \texttt{Computer\_terminal} tool rather than model-action steps, leaving 124 AG trajectories and all 58 HC trajectories for evaluation.

\subsection{Who\&When Pro}
\label{app:benchmark-whowhen-pro}

Who\&When Pro~\citep{liu2026whowhenpro} is a large-scale failure-attribution benchmark containing $12{,}326$ failed trajectories drawn from 26 source benchmarks, nine task categories, and 15 agent frameworks. It covers text, image, and video modalities as well as both single-agent and multi-agent execution settings. The benchmark defines 18 error modes spanning perception, reasoning, planning, action, verification, and coordination failures, and provides gold-standard labels for the responsible agent, the decisive error step, and the corresponding error mode.

Rather than inferring failure causes retrospectively, Who\&When Pro constructs each failed trajectory by injecting a controlled error into a successful execution and retaining only interventions that change the final outcome from success to failure, providing causally grounded labels for the responsible agent and decisive step. Our experiments use all $6{,}257$ text trajectories, which span 16 source benchmarks, eight agent frameworks, and 14 represented error modes; Table~\ref{tab:whowhen-pro-text-composition} summarizes their category-level composition and trajectory lengths.

To keep trajectories from the same task together, we group them by benchmark, task query, and reference answer and assign entire groups to an 8:2 training--test split using seed 42. The partitions contain $5{,}006$ and $1{,}251$ trajectories, respectively, with no shared task groups or normalized queries. Since the training partition contains only failed trajectories, we use $1{,}044$ successful trajectories from ReCast-2K to compute the deviation reference mean and provide contrastive negatives. 

\begin{table}[t]
\centering
\caption{\textbf{Composition of the Who\&When Pro text-modality subset used in our experiments.} Step counts refer to aligned model-action steps, and token counts are computed over the corresponding inputs.}
\label{tab:whowhen-pro-text-composition}
\small
\setlength{\tabcolsep}{5pt}
\renewcommand{\arraystretch}{1.2}
\begin{tabular*}{\textwidth}{@{\extracolsep{\fill}}lrrrrrrr@{}}
\hline
\multirow{2}{*}{\textbf{Task Category}} & \multicolumn{3}{c}{\textbf{\# Trajectories}} & \multicolumn{2}{c}{\textbf{\# Steps}} & \multicolumn{2}{c}{\textbf{\# Tokens}} \\
\cline{2-4}\cline{5-6}\cline{7-8}
& \textbf{Total} & \textbf{Train} & \textbf{Test} & \textbf{Avg.} & \textbf{Max} & \textbf{Avg.} & \textbf{Max} \\
\hline
Embodied & 311 & 249 & 62 & 30.0 & 30 & 1,444.6 & 1,636 \\
STEM & 1,517 & 1,213 & 304 & 5.6 & 30 & 4,199.5 & 23,786 \\
Data Science & 2,446 & 1,958 & 488 & 3.7 & 17 & 609.1 & 6,861 \\
Deep Search & 1,641 & 1,313 & 328 & 5.6 & 31 & 1,551.9 & 21,948 \\
Coding & 342 & 273 & 69 & 3.0 & 3 & 3,127.6 & 7,490 \\
\hline
Total & 6,257 & 5,006 & 1,251 & 5.9 & 31 & 1,906.0 & 23,786 \\
\hline
\end{tabular*}
\end{table}

\subsection{AFTraj-2K}
\label{app:benchmark-aftraj}

AFTraj-2K~\citep{zhang2026agentforesight} contains $2{,}276$ multi-agent trajectories across the Coding, Math, and Agentic domains, including $1{,}162$ successful trajectories and $1{,}114$ failed trajectories with decisive-error annotations. Each domain pairs representative task benchmarks with suitable multi-agent frameworks. The Coding domain uses HumanEval+ and MBPP+~\citep{liu2023evalplus}, with trajectories generated by AutoGen~\citep{wu2024autogen} and MetaGPT~\citep{hong2024metagpt}. The Math domain uses MATH-500~\citep{hendrycks2021math} with AutoGen, while the Agentic domain uses GAIA~\citep{mialon2024gaia} and HotpotQA~\citep{yang2018hotpotqa} with Smolagents~\citep{roucher2025smolagents}. Across the three domains, trajectories contain 11.5 turns on average.

Within AFTraj-2K, successful trajectories are selected through strict checks for outcome correctness, execution integrity, and step coherence. Failed trajectories come from two sources: controlled error injections and naturally failed runs. For injected failures, the modified agent and step serve as direct labels; for natural failures, the earliest decisive error is identified through multi-LLM proposal and verification. The dataset is divided into $1{,}944$ training and 332 test trajectories, with each successful trajectory and its injected variants kept in the same partition. For the in-domain setting, we train on the AFTraj-2K training split and evaluate on its test split.

\subsection{TraceElephant}
\label{app:benchmark-traceelephant}

TraceElephant~\citep{chen2026traceElephant} is a failure-attribution benchmark with fully observable execution traces and reproducible environments. For every agent action, it records the input context, model output, and associated tool interactions, together with trace-level information such as the task instruction, agent configuration, and system architecture. Our evaluation uses two multi-agent system subsets. Captain-Agent~\citep{song2025captainagent} dynamically assembles task-specific teams and contributes 85 failed trajectories, including 73 from GAIA~\citep{mialon2024gaia} and 12 from AssistantBench~\citep{yoran2024assistantbench}. Magentic-One~\citep{fourney2024magentic} uses a fixed multi-agent architecture; after excluding one AssistantBench trajectory lacking a root-cause label, its evaluation subset contains 90 failed trajectories, including 74 from GAIA and 16 from AssistantBench. Together, these two subsets contain 175 failed trajectories, with an average of 25.1 and a maximum of 59 candidate steps per trajectory. None of these trajectories exactly overlaps with the failed trajectories in our training corpus.

\subsection{Data Processing}
\label{app:benchmark-data-processing}

To standardize the heterogeneous trace formats across benchmarks, we use the recorded step roles to retain model-generated actions as attribution candidates and exclude environment observations, tool-result messages, and other non-model events. Among the failed AFTraj-2K test trajectories, this processing reduces the pooled candidate population from 1,711 to 1,548 steps while retaining all 163 trajectories and their annotated root-cause steps. All compared feature-based methods use the same processed candidate set.

% !TEX root = ../../main.tex
\section{Training Dataset Construction}
\label{app:dataset-construction}

\begin{table}[t]
\centering
\caption{\textbf{Composition of our source training dataset.} Initial runs are the aligned executions before filtering. The final total consists of retained successful trajectories and failed trajectories with root-cause annotations. Additional generation runs are grouped under Others.}
\label{tab:source-training-composition}
\small
\setlength{\tabcolsep}{4pt}
\setlength{\aboverulesep}{0pt}
\setlength{\belowrulesep}{0pt}
\renewcommand{\arraystretch}{1.25}
\begin{tabular*}{\textwidth}{@{\extracolsep{\fill}}lllrrrr@{}}
\hline
\multirow{2}{*}{\textbf{Benchmark}} & \multirow{2}{*}{\textbf{MAS}} & \multirow{2}{*}{\textbf{Model}} & \multicolumn{1}{c}{\textbf{Initial}} & \multicolumn{3}{c}{\textbf{Final Dataset}} \\
\cmidrule(lr){4-4}\cmidrule(lr){5-7}
& & & \textbf{Runs} & \textbf{Succ.} & \textbf{Fail.} & \textbf{Total} \\
\hline
\multirow{4}{*}{AssistantBench} & \multirow{4}{*}{Captain-Agent} & DeepSeek V4 Flash & 165 & 20 & 53 & 73 \\
& & GPT-4o-0806 & 165 & 14 & 38 & 52 \\
& & Qwen3-235B & 165 & 20 & 53 & 73 \\
& & \textit{Subtotal} & 495 & 54 & 144 & 198 \\
\hline
\multirow{9}{*}{GAIA} & \multirow{5}{*}{Captain-Agent} & DeepSeek V4 Flash & 451 & 257 & 189 & 446 \\
& & GPT-4o-0806 & 421 & 79 & 186 & 265 \\
& & Qwen3-235B & 449 & 160 & 195 & 355 \\
& & Others & 41 & 6 & 0 & 6 \\
& & \textit{Subtotal} & 1,362 & 502 & 570 & 1,072 \\
\cline{2-7}
& \multirow{4}{*}{Magentic-One} & DeepSeek V4 Flash & 449 & 177 & 258 & 435 \\
& & GPT-4o-0806 & 460 & 132 & 291 & 423 \\
& & Qwen3-235B & 456 & 179 & 233 & 412 \\
& & \textit{Subtotal} & 1,365 & 488 & 782 & 1,270 \\
\hline
\textbf{Total} & -- & -- & \textbf{3,222} & \textbf{1,044} & \textbf{1,496} & \textbf{2,540} \\
\hline
\end{tabular*}
\end{table}

\subsection{Task Sampling and Trajectory Generation}
\label{app:dataset-task-generation}

We construct ReCast-2K using the TraceElephant data-collection pipeline~\citep{chen2026traceElephant}. Captain-Agent~\citep{song2025captainagent} trajectories are collected on GAIA~\citep{mialon2024gaia} and AssistantBench~\citep{yoran2024assistantbench}, and Magentic-One~\citep{fourney2024magentic} trajectories on GAIA. Tasks are drawn from the validation splits. Executions primarily use DeepSeek V4 Flash~\citep{xu2026deepseek}, GPT-4o-0806~\citep{hurst2024gpt}, and Qwen3-235B~\citep{qwen3technicalreport}. The collection contains $3{,}222$ trajectories before filtering, as summarized in Table~\ref{tab:source-training-composition}.

\paragraph{Captain-Agent generation.}
Captain-Agent dynamically constructs a task-specific team for every execution. Its builder is instructed to include capabilities for information retrieval, Python-based file processing and calculation, and answer verification, and selects concrete experts from the Captain-Agent agent library. Each run is allowed at most 30 group-chat rounds. Generation uses temperature 0.1, top-$p$ 0.95, and a maximum of 4,096 output tokens per model call; failed or empty executions are retried up to five times.

\paragraph{Magentic-One generation.}
Magentic-One uses its fixed centrally orchestrated team: an Orchestrator maintains the plan and delegates to WebSurfer, FileSurfer, Coder, and ComputerTerminal. WebSurfer operates a browser, FileSurfer reads local artifacts, Coder proposes programs, and ComputerTerminal executes them in an isolated run directory. As with Captain-Agent, attached GAIA files are exposed through their resolved local paths, and each execution is capped at 30 rounds. 

\subsection{Root-Cause Annotation}
\label{app:dataset-root-cause-annotation}

Before annotation, we filter incomplete or malformed traces, normalize agent identities and metadata, and extract the final answer from each run. We determine execution outcomes by matching the extracted answers against benchmark references using both exact and semantic criteria. Failure preprocessing removes traces without a usable answer or with at most one step and applies answer-based deduplication within each task and source configuration. Table~\ref{tab:source-training-composition} reports the final retained trajectories.

For each retained failed trajectory, we apply an all-at-once annotation protocol using three independent LLM judges: GPT-5.5~\citep{openai2026gpt55}, Gemini 3 Pro Preview~\citep{google2025gemini3}, and Claude Opus 4.6~\citep{anthropic2026opus46}. Each judge predicts a root-cause step, and we aggregate valid returned predictions by step-level voting. In the final corpus, 1,112 trajectories carry consensus labels: 766 have unanimous agreement among valid predictions, and 346 have a two-of-three majority. The remaining 384 trajectories undergo manual review to determine their final root-cause labels. Together, these groups account for all 1,496 failed trajectories. The responsible agent is the agent acting at the root-cause step.

\subsection{Final Dataset Statistics}
\label{app:dataset-statistics}

Table~\ref{tab:source-training-statistics} summarizes the final dataset by generation model and trajectory type. It contains $1{,}044$ successful and $1{,}496$ failed trajectories with substantial variation in step and token counts. Two patterns emerge from the collected data: \textbf{(1) Failed trajectories are generally longer than successful ones.} They average 25.9 versus 17.2 steps and $8{,}346.6$ versus $5{,}605.5$ tokens. \textbf{(2) Trajectory characteristics vary noticeably across generation models.} GPT-4o-0806 produces failed trajectories with the highest average number of steps (32.0), whereas DeepSeek V4 Flash yields the highest average and maximum token counts among failed trajectories. Together, these variations provide broad coverage of trajectory lengths and execution patterns in the training data.

\begin{table}[t]
\centering
\caption{\textbf{Statistics of ReCast-2K.} Steps count recorded model responses, while token counts cover the question prefix and serialized trajectory using the Qwen3.5-27B tokenizer. Overall averages are computed over all trajectories.}
\label{tab:source-training-statistics}
\small
\setlength{\tabcolsep}{5pt}
\renewcommand{\arraystretch}{1.25}
\begin{tabular*}{\textwidth}{@{\extracolsep{\fill}}llrrrrr@{}}
\hline
\multirow{2}{*}{\textbf{Model}} & \multirow{2}{*}{\textbf{Type}} & \multirow{2}{*}{\textbf{\# Trajectories}} & \multicolumn{2}{c}{\textbf{\# Steps}} & \multicolumn{2}{c}{\textbf{\# Tokens}} \\
\cline{4-5}\cline{6-7}
& & & \textbf{Avg.} & \textbf{Max} & \textbf{Avg.} & \textbf{Max} \\
\hline
\multirow{2}{*}{DeepSeek V4 Flash} & Successful & 454 & 15.7 & 61 & 7,005.5 & 59,153 \\
& Failed & 500 & 21.6 & 55 & 10,603.5 & 74,773 \\
\hline
\multirow{2}{*}{GPT-4o-0806} & Successful & 225 & 18.5 & 60 & 4,460.0 & 22,757 \\
& Failed & 515 & 32.0 & 61 & 7,996.1 & 29,094 \\
\hline
\multirow{2}{*}{Qwen3-235B} & Successful & 359 & 18.5 & 62 & 4,568.5 & 21,378 \\
& Failed & 481 & 24.0 & 62 & 6,375.8 & 23,621 \\
\hline
Others & Successful & 6 & 4.8 & 7 & 4,677.5 & 6,729 \\
\hline
\multirow{2}{*}{\textbf{Overall}} & Successful & 1,044 & 17.2 & 62 & 5,605.5 & 59,153 \\
& Failed & 1,496 & 25.9 & 62 & 8,346.6 & 74,773 \\
\hline
\end{tabular*}
\end{table}

% !TEX root = ../../main.tex
\section{Evaluation Details and Hyperparameters}
\label{app:evaluation-details}

\subsection{Evaluation Metrics}
\label{app:evaluation-metrics}

Using the root-cause timestamp $t_i^*$ defined in Section~\ref{sec:problem-definition}, let $\operatorname{rank}_i(t_i^*)$ denote the predicted rank of the root cause in failed trajectory $\tau_i$. Hit@$k$ and MRR over the evaluated failed trajectories are computed as
\[
\operatorname{Hit@}k=\frac{1}{|\mathcal{T}^{\mathrm{fail}}|}\sum_{\tau_i\in\mathcal{T}^{\mathrm{fail}}}\mathbb{I}\!\left[\operatorname{rank}_i(t_i^*)\leq k\right],
\qquad
\operatorname{MRR}=\frac{1}{|\mathcal{T}^{\mathrm{fail}}|}\sum_{\tau_i\in\mathcal{T}^{\mathrm{fail}}}\frac{1}{\operatorname{rank}_i(t_i^*)}.
\]
We report Hit@$k$ for $k\in\{1,3,5\}$. Results are reported as mean $\pm$ population standard deviation over three seeds.

For score-based evaluation, we use the attribution score $\alpha_{i,t}$ assigned to timestamp $t$. The root-cause step $t_i^*$ is the sole positive candidate; all timestamps in $\mathcal{N}(\tau_i)$ are negative candidates. Their total numbers over the evaluated failed trajectories are
\[
C_{+}=|\mathcal{T}^{\mathrm{fail}}|,
\qquad
C_{-}=\sum_{\tau_i\in\mathcal{T}^{\mathrm{fail}}}|\mathcal{N}(\tau_i)|.
\]
AUROC and AUPRC pool these candidate-level predictions across trajectories rather than averaging per-trajectory values. AUROC compares every annotated root-cause step with every non-root step across the evaluated trajectories:
\begin{equation}
\label{eq:evaluation-auroc}
\operatorname{AUROC}
=
\frac{1}{C_{+}C_{-}}
\sum_{\tau_i\in\mathcal{T}^{\mathrm{fail}}}
\sum_{\tau_j\in\mathcal{T}^{\mathrm{fail}}}
\sum_{t\in\mathcal{N}(\tau_j)}
\left[
\mathbb{I}(\alpha_{i,t_i^*}>\alpha_{j,t})
+
\frac{1}{2}\mathbb{I}(\alpha_{i,t_i^*}=\alpha_{j,t})
\right].
\end{equation}

For AUPRC, we aggregate the attribution scores of all candidate steps across the evaluated trajectories and sort the $J$ distinct scores in descending order as $\theta_1>\theta_2>\cdots>\theta_J$. At threshold $\theta_j$, every candidate step with $\alpha_{i,t}\geq\theta_j$ is treated as a positive prediction. Such a prediction is a true positive if $t=t_i^*$ and a false positive otherwise. The corresponding counts are
\[
\operatorname{TP}_{j}
=
\sum_{\tau_i\in\mathcal{T}^{\mathrm{fail}}}
\mathbb{I}(\alpha_{i,t_i^*}\geq\theta_j),
\qquad
\operatorname{FP}_{j}
=
\sum_{\tau_i\in\mathcal{T}^{\mathrm{fail}}}
\sum_{t\in\mathcal{N}(\tau_i)}
\mathbb{I}(\alpha_{i,t}\geq\theta_j).
\]
Precision and recall are
\[
\operatorname{Prec}_{j}
=
\frac{\operatorname{TP}_{j}}
{\operatorname{TP}_{j}+\operatorname{FP}_{j}},
\qquad
\operatorname{Rec}_{j}
=
\frac{\operatorname{TP}_{j}}{C_{+}}.
\]
We compute AUPRC using the non-interpolated average-precision formulation:
\begin{equation}
\label{eq:evaluation-auprc}
\operatorname{AUPRC}
=
\sum_{j=1}^{J}
\left(
\operatorname{Rec}_{j}-\operatorname{Rec}_{j-1}
\right)
\operatorname{Prec}_{j},
\qquad
\operatorname{Rec}_{0}=0.
\end{equation}

\subsection{Hyperparameter Settings}
\label{app:hyperparameter-settings}

\begin{table}[t]
\centering
\small
\caption{Hyperparameter settings for our method.}
\label{tab:hyperparameter-settings}
\setlength{\tabcolsep}{5pt}
\renewcommand{\arraystretch}{1.08}
\begin{tabular*}{\textwidth}{@{\extracolsep{\fill}}lll@{}}
\toprule
\textbf{Component} & \textbf{Hyperparameter} & \textbf{Value} \\
\midrule
Backbone & Hidden-state dimension $d_h$ & 5,120 \\
\midrule
Layer selection & Cross-validation folds & 5 \\
& Layer budget $k$ & 8 \\
& Ridge ratio & 0.10 \\
& Selected layers & $\{7,15,19,32,33,47,49,57\}$ \\
\midrule
Projection & Dimension per branch $d_p$ & 2,048 \\
& Final step-feature dimension $2d_p$ & 4,096 \\
& Random state & 42 \\
\midrule
Encoder & Transformer layers / attention heads & 4 / 4 \\
& Hidden dimension & 128 \\
& Output dimension $d_z$ & 128 \\
& Feed-forward dimension & 384 \\
& Maximum sequence length (steps) & 512 \\
& Dropout & 0.10 \\
\midrule
Augmentation & Step / feature masking probability & 0.10 / 0.10 \\
& Gaussian-noise standard deviation & 0.01 \\
\midrule
Objective & InfoNCE temperature $\kappa$ & 0.12 \\
& $(\lambda_{\mathrm{rank}},\lambda_{\mathrm{root}},\lambda_{\mathrm{non\text{-}root}})$ & $(1.0,0.95,0.25)$ \\
\midrule
Optimization & Optimizer & AdamW \\
& Learning rate / weight decay & $2\times10^{-4}$ / $10^{-4}$ \\
& Batch size & 32 \\
& Maximum epochs & 50 \\
& Gradient clipping & 1.0 \\
& Training seeds & 6, 20, 42 \\
\bottomrule
\end{tabular*}
\end{table}

Table~\ref{tab:hyperparameter-settings} summarizes the settings used with Qwen3.5-27B~\citep{qwen3.5} as the hidden-state extraction backbone. The selected eight layers provide features for the pattern and deviation branches, each with dimension $d_p=2{,}048$. A four-layer bidirectional Transformer maps their concatenation to $d_z=128$ step representations. The backbone, projection matrix, and successful-reference mean remain fixed while the encoder and scoring head are jointly optimized. 

% !TEX root = ../../main.tex

\section{Baseline Settings}
\label{app:baseline-settings}

\subsection{Heuristic Baselines}
\label{app:baseline-basic}

\paragraph{First.}
First is a deterministic position-only baseline that ranks candidate steps by their step order and selects the earliest step as the predicted root cause. It has no trainable parameters and does not use the trajectory text, agent identity, or benchmark labels. Its standard deviation across seeds is zero because the ranking is deterministic.

\paragraph{Random.}
Random produces a complete ranking by uniformly shuffling all candidate steps. For a fair comparison, we evaluate it with the same three seeds as the other methods and report the mean and population standard deviation. Both First and Random operate directly on each benchmark's original candidate set.

\subsection{LLM-Based Methods}
\label{app:baseline-llm}

\paragraph{Common protocol.}
We adopt two standard LLM-based attribution protocols from Who\&When~\citep{zhang2025whowhen}. Appendix~\ref{app:baseline-prompts} provides the complete prompts used in all evaluations, including the task instruction, \textbf{ground-truth answer}, and serialized multi-agent conversation. Each run returns at most one predicted step. A prediction is counted as correct only when the parsed step maps to a valid candidate and matches the annotated root-cause step exactly; an unparsable response, an out-of-candidate step, or a run that finds no error is counted as incorrect. Since these protocols return a single step rather than a complete ranking, we report only Hit@1. The two protocols operate as follows:
\begin{itemize}[leftmargin=1em]
\item \textbf{All-at-Once.} The model receives the complete conversation in one request and is asked to return the responsible agent, the first erroneous step, and a short explanation.
\item \textbf{Step-by-Step.} The model processes growing trajectory prefixes in step order. At each step, the prompt asks whether the most recent action contains an error that could derail the execution. Evaluation stops at the first response parsed as ``Yes,'' and that step becomes the prediction. If all steps receive ``No'' or no valid decision is produced, the trajectory is marked as having no predicted root cause. This protocol uses more calls than All-at-Once but prevents later trajectory content from influencing an earlier decision.
\end{itemize}

\paragraph{Models and inference settings.}
We evaluate both protocols with Qwen3.5-27B~\citep{qwen3.5}, DeepSeek V4 Flash~\citep{xu2026deepseek}, and GPT-5.6 Luna~\citep{openai2026gpt56}. Qwen3.5-27B is served locally with vLLM using bfloat16 weights, greedy decoding with temperature zero, thinking disabled, and a configured 131K-token context window. Across all four benchmarks, all three models use a maximum output length of 16,384 tokens under both protocols. DeepSeek V4 Flash uses explicitly enabled thinking, while GPT-5.6 Luna retains its default configuration. Both protocols are evaluated with each of the three models using inference seeds $\{6,20,42\}$, and results are reported as mean $\pm$ population standard deviation.

\paragraph{AgenTracer.}
AgenTracer~\citep{zhang2026agentracer} constructs attribution training data through counterfactual replay and fault injection and trains an 8B attribution model with reinforcement learning. Since the model is not publicly available, we directly use its reported Who\&When results, without re-evaluation on our filtered Algorithm subset. These are single reported values and do not have a three-seed standard deviation.

\paragraph{CHIEF.}
CHIEF~\citep{wang2026chief} constructs hierarchical causal graphs for failure attribution. We follow its released six-stage pipeline, retaining the original prompts, parsers, and retrieval setup. Inference uses DeepSeek V4 Flash with thinking enabled, a maximum output length of 16,384 tokens per call, and seeds $\{6,20,42\}$. Retrieval uses the supplied knowledge bases and all-MiniLM-L6-v2 embeddings. Following the original protocol, prompts include task reference answers but not root-cause annotations. We evaluate the final predicted step using Hit@1.

\subsection{Feature-Based Methods}
\label{app:baseline-small-models}

\begin{table}[t]
\centering
\small
\caption{\textbf{Optimization settings for trainable feature-based baselines.}}
\label{tab:small-model-optimization}
\renewcommand{\arraystretch}{1.25}
\begin{tabular}{lcccl}
\hline
\textbf{Method} & \textbf{LR} & \textbf{Weight decay} & \textbf{Batch} &
\textbf{Budget} \\
\hline
OAT & $10^{-4}$ & $10^{-5}$ & 32 & 300 epochs \\
StepFinder & $10^{-3}$ & $10^{-5}$ & 16 & 50 epochs \\
ASCon & $5{\times}10^{-5}$ & $10^{-5}$ & 1 & 12 epochs \\
\hline
\end{tabular}
\end{table}

\paragraph{OAT.}
OAT~\citep{yeh2026oat} formulates unsupervised failure attribution as one-class learning: it learns the normal execution dynamics of successful trajectories and treats deviations from these dynamics as failure signals. We extract the query and candidate-action representations from the final layer of Qwen3.5-27B, using mean pooling over the corresponding tokens. For each seed, the successful trajectories are randomly divided into 80\% training and 20\% validation partitions. PCA is fitted only on the training partition to reduce the 5,120-dimensional representations to 64 dimensions; the projected features are then standardized by subtracting the training-partition mean and dividing by its standard deviation. The query representation initializes the trajectory dynamics, and step timestamps are normalized to $[0,1]$.

The predictor models each successful trajectory with a Neural CDE whose projected input and latent-state dimensions are both 64. We directly use the released OAT implementation, which employs a three-layer MLP vector field, cubic interpolation, the Euler solver, and a four-layer control gate with hidden width 12. The query-conditioned initializer and hidden-state bridge are three-layer MLPs with hidden width 64; adjoint computation and initial-state norm regularization are disabled. We optimize the model with AdamW using the settings in Table~\ref{tab:small-model-optimization}. At evaluation time, the squared reconstruction error $e_t=\lVert\mathbf{h}_t-\widehat{\mathbf{h}}_t\rVert_2^2$ is used as the anomaly score for step $t$, and steps with larger scores are ranked earlier.

\paragraph{StepFinder.}
StepFinder~\citep{zhu2026stepfinder} is a supervised lightweight attribution model that represents a trajectory as parallel sequences of step-content and agent-identity embeddings. We separately encode the content and agent name of each candidate step with Qwen3-Embedding-0.6B~\citep{zhang2025qwen3embedding}, using last-token pooling and a maximum input length of 8,192 tokens, and retain 128 and 32 dimensions, respectively. A two-layer BiLSTM with hidden size 64 per direction and dropout 0.5 captures the evolution of the trajectory. The resulting step representations are processed by a two-head agent-aware interaction module with head dimension 32. This module augments global step attention with an agent-similarity bias and a trajectory-level agent gate, both controlled by $\alpha=0.1$.

StepFinder first predicts a learned error logit for each step and then refines it using normalized differences between steps at scales $\{1,2\}$ and a linearly decaying position bias that favors earlier root causes. We set the corresponding weights to $\beta=0.9$ and $\gamma=0.4$. Training minimizes trajectory-level cross-entropy over the candidate steps together with an auxiliary loss for predicting the next-step representation; the auxiliary loss weight is $\lambda=0.9$. We use AdamW, gradient clipping at 1.0, and the optimization settings in Table~\ref{tab:small-model-optimization}. The same configuration is used across all reported benchmarks. Candidate steps are ranked by the resulting refined error scores.

\paragraph{MASPrism.}
MASPrism~\citep{liu2026masprism} is a training-free and oracle-free attribution method that extracts diagnostic signals during two prefill passes of a frozen language model. We follow its original two-stage scoring protocol without training an attribution head or using root-cause annotations. The Filtering stage uses step-level negative log-likelihood to identify the top 20\% downstream symptom steps and attention from these symptoms to retain up to five earlier candidate sources. The Diagnosis stage restores the selected symptom and candidate contents, recomputes the prefill signals, and combines normalized attention, NLL contrast, and multi-symptom consensus to rank the candidate steps. We use a consensus top-$5$ cutoff and set the consensus coefficient to 0.3.

We use the frozen Qwen3-0.6B~\citep{qwen3technicalreport} in bfloat16. Each trajectory is serialized with the model's chat template without the ground-truth answer, and the maximum input length is set to 16,384 tokens. Following the original layer-selection rule, we average attention over all heads and the final 20\% of the 28 Transformer blocks, corresponding to layers 22--27 under zero-based indexing. The Filtering stage retains up to 512 prefix tokens per step. During Diagnosis, the selected symptom and candidate steps are restored in full, while each remaining ordinary step retains up to 128 prefix tokens. These larger input budgets reduce truncation as a potential confound. The method performs only two prefill passes, requires no decoding, and is deterministic; the three reported seeds therefore produce identical predictions.

\paragraph{ASCon.}
ASCon~\citep{jiang2026ascon} jointly contextualizes step and agent representations to support failure attribution. We use its step-level predictions under the failure attribution setting to localize the root-cause step. In our implementation, candidate-step text and each unique agent name are encoded separately with Qwen3.5-27B~\citep{qwen3.5} using last-token pooling to obtain 5,120-dimensional embeddings. Feature extraction uses bfloat16 weights, a maximum input length of 1,024 tokens, and an embedding batch size of 4. Consecutive candidate steps form a directed step graph, while transitions between their recorded actors form a directed agent graph. Under our unified candidate representation, the graphs contain only the retained candidate-step nodes and their corresponding actor nodes; the user query and user role are not instantiated as additional graph nodes. A two-layer direction-aware graph attention network separately aggregates preceding, succeeding, and self information to obtain contextualized step representations.

ASCon next aggregates the contextualized representations of each agent's steps through masked step-to-agent attention, producing a behavior-aware agent representation. It then incorporates the agent identity and global trajectory context and propagates information over the directed agent graph. The resulting representation of each agent is fused back into the representations of its steps, allowing step-level predictions to incorporate agent-specific behavioral context. We use two direction-aware graph attention layers, 256-dimensional step representations, 128-dimensional agent representations, a 128-dimensional step-to-agent attention space, and dropout 0.1. Separate step and agent classifiers are trained jointly with class-balanced binary cross-entropy using loss weights $\lambda_{\mathrm{step}}=0.3$ and $\lambda_{\mathrm{agent}}=1.0$. Optimization uses AdamW for 12 epochs with gradient clipping at 1.0 and the remaining settings in Table~\ref{tab:small-model-optimization}. Candidate steps are ranked by the step-classifier logits.

% !TEX root = ../../main.tex

\section{Additional Experimental Results}
\label{app:additional-experiments}

This section collects additional experimental results omitted from the main text, including complete evaluation metrics, detailed subset analyses, backbone comparisons, and time efficiency analysis.

\subsection{MRR Across Rank Cutoffs}
\label{app:additional-mrr}

\begin{figure}[!ht]
\centering
\includegraphics[width=\textwidth]{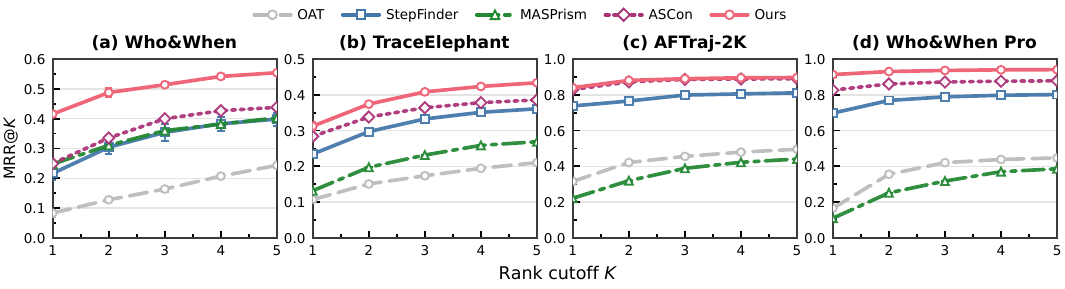}
\caption{MRR@$K$ across rank cutoffs $K=1,\ldots,5$ on the four benchmarks. Curves show the cumulative reciprocal-rank score as the cutoff increases; error bars denote the population standard deviation over three seeds.}
\label{fig:whowhen-traceelephant-mrrk}
\end{figure}

MRR@$K$ complements Hit@$K$ by assigning a reciprocal-rank score to the root cause when it appears within the top $K$ positions and zero otherwise, rewarding methods that place the root cause earlier within a fixed inspection budget. Figure~\ref{fig:whowhen-traceelephant-mrrk} reports this metric for rank cutoffs from 1 to 5. ReCast achieves the highest MRR@$K$ among the compared methods at every cutoff across all four benchmarks. On TraceElephant, ReCast reaches MRR@1 of 0.3124 and MRR@5 of 0.4341, compared with 0.2838 and 0.3860 for ASCon.

\subsection{Score-Level Discrimination}
\label{app:additional-score-discrimination}

\begin{figure}[!ht]
\centering
\includegraphics[width=\textwidth]{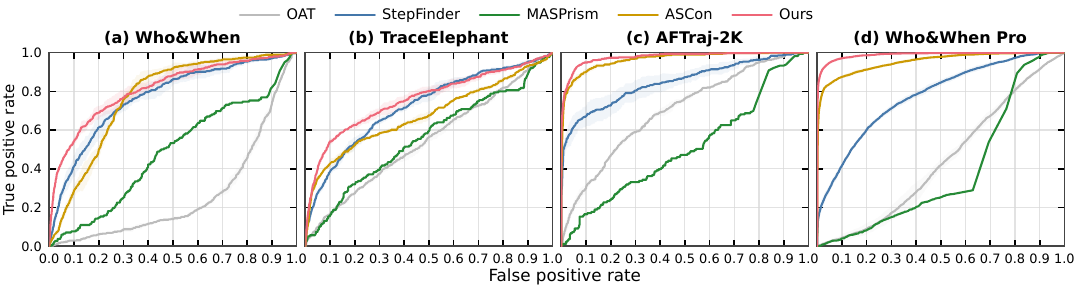}
\caption{Receiver operating characteristic (ROC) curves on the four benchmarks. Each curve is computed by pooling the candidate-step predictions within a benchmark. Solid lines show the mean over three seeds, and shaded regions indicate the population standard deviation.}
\label{fig:four-benchmark-roc}
\end{figure}

\begin{figure}[!ht]
\centering
\includegraphics[width=\textwidth]{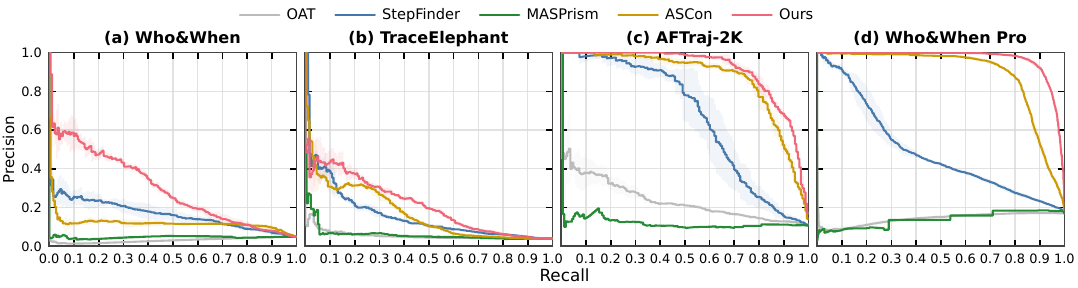}
\caption{Precision--recall (PR) curves on the four benchmarks. Each curve is computed by pooling the candidate-step predictions within a benchmark. Solid lines show the mean over three seeds, and shaded regions indicate the population standard deviation.}
\label{fig:four-benchmark-pr}
\end{figure}

The rank-based results above measure whether each trajectory's root cause is prioritized over the other steps in the same trajectory. We additionally evaluate whether the raw attribution scores consistently separate root-cause steps from non-root steps after pooling all candidate steps within each benchmark's evaluation population. Figures~\ref{fig:four-benchmark-roc} and~\ref{fig:four-benchmark-pr} show the resulting ROC and PR curves, respectively. The corresponding AUROC and AUPRC follow the definitions in Appendix~\ref{app:evaluation-metrics}.

ReCast achieves the highest AUROC and AUPRC on all four benchmarks. Its advantage in AUPRC is particularly meaningful under class imbalance, as PR curves capture the trade-off between recovering root-cause steps and avoiding false positives. On Who\&When and TraceElephant, it obtains AUROC values of $0.8139$ and $0.7583$ and AUPRC values of $0.2964$ and $0.2125$, respectively. On AFTraj-2K and Who\&When Pro, ReCast reaches AUROC values of $0.9775$ and $0.9887$ together with AUPRC values of $0.9082$ and $0.9667$.

Together, the MRR@$K$, AUROC, and AUPRC results support the effectiveness of ReCast for failure attribution. Its advantage extends from prioritizing the root-cause step within each trajectory to distinguishing root-cause steps from other candidates across trajectories.

\subsection{Hidden-State Backbone Comparison}
\label{app:additional-backbone-comparison}

\begin{table}[!ht]
\centering
\small
\caption{Hidden-state backbone configurations. Each backbone retains eight layers selected independently using the same source-only procedure.}
\label{tab:backbone-configurations}
\setlength{\tabcolsep}{6pt}
\renewcommand{\arraystretch}{1.12}
\begin{tabular}{lrrl}
\toprule
\textbf{Backbone} & \textbf{\# Layers} & \textbf{Hidden size} & \textbf{Selected layers} \\
\midrule
Qwen3.5-0.8B & 24 & 1,024 & 3, 5, 8, 11, 13, 18, 19, 22 \\
Qwen3.5-9B   & 32 & 4,096 & 3, 7, 11, 16, 17, 23, 26, 31 \\
Qwen3.5-27B  & 64 & 5,120 & 7, 15, 19, 32, 33, 47, 49, 57 \\
\bottomrule
\end{tabular}
\end{table}

\paragraph{Protocol.}
To examine whether attribution performance scales monotonically with backbone size, we compare Qwen3.5 models with 0.8B, 9B, and 27B parameters while holding the downstream ReCast pipeline fixed. For each backbone, we independently select eight layers using attribution-aware layer selection on ReCast-2K, set the projection dimension of each branch to $d_p=2{,}048$, and retain the same encoder architecture. Training and evaluation follow the same settings as the main experiments. Results are reported as mean $\pm$ population standard deviation over seeds 6, 20, and 42. Table~\ref{tab:backbone-configurations} summarizes the backbone configurations and selected layers.

\begin{table}[!ht]
\centering
\scriptsize
\caption{Effect of the hidden-state extraction backbone. Results report mean $\pm$ population standard deviation over three encoder seeds. Bold indicates the best mean within each evaluation population, determined before rounding.}
\label{tab:backbone-complete-results}
\setlength{\tabcolsep}{2.5pt}
\renewcommand{\arraystretch}{1.10}
\resizebox{\textwidth}{!}{%
\begin{tabular}{llrrrrrr}
\toprule
\textbf{Evaluation population} & \textbf{Backbone} &
\textbf{Hit@1 (\%)} & \textbf{Hit@3 (\%)} & \textbf{Hit@5 (\%)} &
\textbf{MRR (\%)} & \textbf{AUROC (\%)} & \textbf{AUPRC (\%)} \\
\midrule
\multirow{3}{*}{\shortstack[l]{Who\&When\\Handcrafted}}
& Qwen3.5-0.8B & 30.46$\pm$2.15 & 41.95$\pm$2.15 & 55.75$\pm$0.81 & 41.80$\pm$1.07 & \textbf{80.00$\pm$2.12} & 19.09$\pm$0.84 \\
& Qwen3.5-9B & 29.89$\pm$3.54 & 44.83$\pm$3.72 & 52.87$\pm$4.30 & 41.61$\pm$0.83 & 79.94$\pm$2.41 & 18.55$\pm$2.91 \\
& Qwen3.5-27B & \textbf{33.33$\pm$1.63} & \textbf{45.98$\pm$0.81} & \textbf{57.47$\pm$0.81} & \textbf{44.21$\pm$0.77} & 79.22$\pm$0.91 & \textbf{21.88$\pm$0.92} \\
\midrule
\multirow{3}{*}{\shortstack[l]{Who\&When\\Algorithm}}
& Qwen3.5-0.8B & 38.17$\pm$2.31 & \textbf{74.73$\pm$2.31} & 91.94$\pm$1.14 & 59.20$\pm$2.00 & 72.00$\pm$0.38 & 33.42$\pm$0.43 \\
& Qwen3.5-9B & 38.98$\pm$0.38 & 74.46$\pm$2.01 & \textbf{92.74$\pm$2.37} & 59.85$\pm$0.67 & 73.64$\pm$1.58 & 35.10$\pm$0.47 \\
& Qwen3.5-27B & \textbf{45.43$\pm$2.01} & 72.31$\pm$1.01 & 92.47$\pm$1.66 & \textbf{63.28$\pm$1.42} & \textbf{74.75$\pm$1.00} & \textbf{38.52$\pm$1.84} \\
\midrule
\multirow{3}{*}{\shortstack[l]{TraceElephant\\Captain-Agent}}
& Qwen3.5-0.8B & \textbf{38.43$\pm$2.42} & \textbf{62.35$\pm$4.19} & 78.82$\pm$2.54 & \textbf{55.12$\pm$2.14} & \textbf{81.07$\pm$2.40} & 26.82$\pm$2.49 \\
& Qwen3.5-9B & 37.25$\pm$2.00 & \textbf{62.35$\pm$2.54} & \textbf{80.00$\pm$1.66} & 54.54$\pm$1.19 & 80.92$\pm$2.03 & \textbf{27.42$\pm$0.37} \\
& Qwen3.5-27B & 35.69$\pm$2.93 & 61.57$\pm$1.47 & 78.04$\pm$2.42 & 52.90$\pm$1.13 & 78.30$\pm$2.00 & 25.30$\pm$0.75 \\
\midrule
\multirow{3}{*}{\shortstack[l]{TraceElephant\\Magentic-One}}
& Qwen3.5-0.8B & 27.78$\pm$0.91 & 42.22$\pm$1.81 & 50.00$\pm$1.57 & 39.81$\pm$0.31 & 71.69$\pm$1.92 & \textbf{18.87$\pm$0.52} \\
& Qwen3.5-9B & \textbf{28.89$\pm$0.91} & 43.70$\pm$2.92 & \textbf{54.81$\pm$3.43} & \textbf{41.07$\pm$1.30} & 71.61$\pm$3.56 & 16.51$\pm$1.76 \\
& Qwen3.5-27B & 27.04$\pm$1.89 & \textbf{46.67$\pm$1.57} & 52.96$\pm$2.10 & 40.52$\pm$1.16 & \textbf{72.78$\pm$0.58} & 17.98$\pm$1.99 \\
\midrule
\multirow{3}{*}{AFTraj-2K}
& Qwen3.5-0.8B & 82.82$\pm$1.81 & 94.68$\pm$0.29 & 97.96$\pm$0.77 & 89.05$\pm$1.06 & 97.63$\pm$0.08 & 88.05$\pm$1.38 \\
& Qwen3.5-9B & 83.64$\pm$1.61 & 94.89$\pm$1.26 & \textbf{98.16$\pm$0.50} & 89.72$\pm$0.95 & \textbf{97.88$\pm$0.24} & 90.54$\pm$0.75 \\
& Qwen3.5-27B & \textbf{84.05$\pm$0.87} & \textbf{95.09$\pm$1.00} & 97.96$\pm$0.29 & \textbf{89.99$\pm$0.49} & 97.75$\pm$0.21 & \textbf{90.82$\pm$0.58} \\
\midrule
\multirow{3}{*}{Who\&When Pro}
& Qwen3.5-0.8B & 87.93$\pm$0.49 & 95.31$\pm$0.14 & 98.13$\pm$0.19 & 92.09$\pm$0.29 & 97.84$\pm$0.11 & 94.25$\pm$0.21 \\
& Qwen3.5-9B & 89.90$\pm$0.04 & 96.16$\pm$0.36 & 98.29$\pm$0.08 & 93.40$\pm$0.09 & 98.35$\pm$0.06 & 95.61$\pm$0.07 \\
& Qwen3.5-27B & \textbf{91.42$\pm$0.21} & \textbf{96.64$\pm$0.26} & \textbf{98.53$\pm$0.14} & \textbf{94.36$\pm$0.11} & \textbf{98.87$\pm$0.12} & \textbf{96.67$\pm$0.17} \\
\bottomrule
\end{tabular}%
}
\end{table}

\paragraph{Results.}
Table~\ref{tab:backbone-complete-results} shows that the benefits of a larger backbone vary across benchmarks and metrics. Qwen3.5-27B achieves the highest Hit@1 on both Who\&When subsets, AFTraj-2K, and Who\&When Pro, with all six metrics improving with backbone size on Who\&When Pro. However, smaller backbones remain competitive: 0.8B and 9B achieve the highest Hit@1 on Captain-Agent and Magentic-One, respectively. On AFTraj-2K, their Hit@1 scores are within 1.23 percentage points of 27B, and 9B achieves the highest Hit@5 and AUROC. These results suggest that larger backbones can improve attribution performance, although the gains vary across benchmarks and metrics. With attribution-aware layer selection and representation learning, smaller backbones can also provide effective signals for failure attribution.

\subsection{Time Efficiency Analysis}
\label{app:time-efficiency}

\begin{figure}[!ht]
\centering
\includegraphics[width=\textwidth]{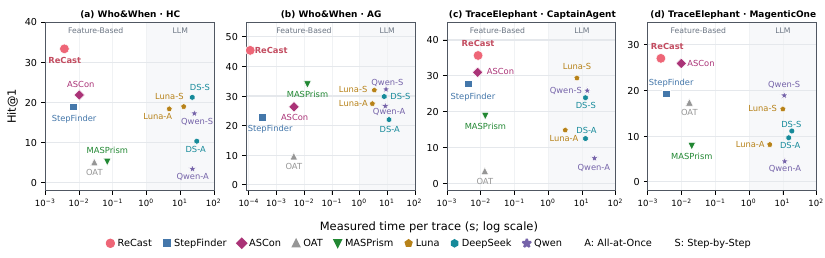}
\caption{Hit@1 accuracy (\%) versus measured time per trajectory on Who\&When and TraceElephant. Points show three-seed means; time is in seconds on a logarithmic scale. Feature-based methods report attribution time excluding feature extraction, while LLM methods report inference time excluding backend initialization. Upper-left positions indicate higher accuracy at lower measured cost. A and S denote All-at-Once and Step-by-Step.}
\label{fig:time-efficiency}
\end{figure}

\paragraph{Protocol.}
We compare Hit@1 accuracy with the average processing time per trajectory on Who\&When and TraceElephant. ReCast, StepFinder, ASCon, and OAT use precomputed features, excluding feature extraction and embedding-cache preparation from attribution time. For MASPrism, we report attribution time excluding model initialization, tokenization, and prefill-stage signal extraction. For LLM-based methods, inference time excludes model initialization but includes prompt preparation and API delays and retries where applicable. We divide the total measured time by the number of evaluated trajectories and report the mean over three seeds in Figure~\ref{fig:time-efficiency}.

\paragraph{Results.}
ReCast achieves the highest Hit@1 across all four subsets with mean attribution times ranging from 0.11 to 8.28 ms per trajectory. It has the lowest mean attribution time among the compared feature-based methods on HC, AG, and Magentic-One. On Captain-Agent, StepFinder and ASCon are faster, but ReCast achieves higher accuracy. In contrast, the LLM-based baselines shown in Figure~\ref{fig:time-efficiency} require 2.88--30.50 seconds per trajectory under the measured configurations while attaining lower Hit@1. This comparison highlights the efficiency of ReCast's attribution stage when step features are available.

% !TEX root = ../../main.tex
\section{Case Study}
\label{app:case-study}

\begin{figure}[t]
\centering
\includegraphics[width=\textwidth]{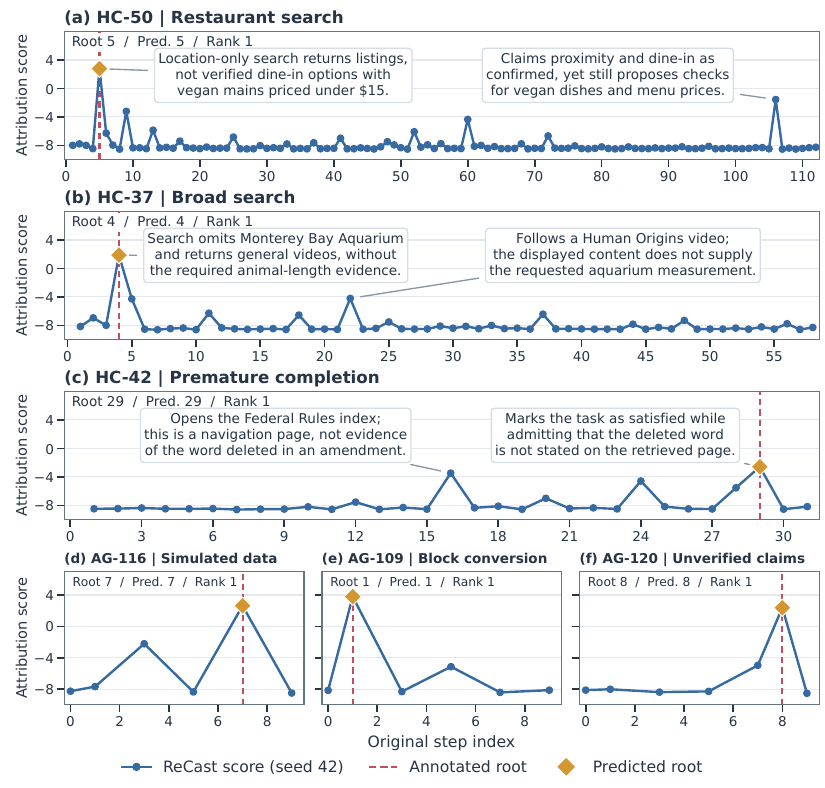}
\caption{Step-level attribution scores on three HC trajectories (a--c) and three AG trajectories (d--f). Original zero-based step indices are retained; gaps indicate events without candidate scores.}
\label{fig:hc-ag-case-study}
\end{figure}

Figure~\ref{fig:hc-ag-case-study} presents six cases from the Handcrafted and Algorithm subsets of Who\&When. The curves show ReCast's attribution scores for all candidate steps, with higher scores indicating higher ranks within each trajectory. In these selected cases, ReCast ranks the annotated root-cause step above other high-scoring candidates, providing qualitative support for the effectiveness of its learned step representations in failure attribution.

\textbf{HC-50, restaurant search (Figure~\ref{fig:hc-ag-case-study}(a)).}
The task asks for dine-in restaurants within one block of Washington Square Park that serve vegan main dishes for under \$15. At step 5, a location-only search returns general listings without verifying the remaining constraints; the dataset annotates this retrieval as the root cause. Later steps also receive elevated scores. For example, the summary at step 106 presents proximity and dine-in service as confirmed while leaving vegan options and prices for further verification. ReCast nevertheless assigns its highest score to step 5, ranking the earlier annotated error above this later competing peak.

\textbf{HC-37, missing a source constraint (Figure~\ref{fig:hc-ag-case-study}(b)).}
The task asks for an organism's maximum body length in meters, as reported by the Monterey Bay Aquarium website. At step 4, the agent searches for National Geographic videos without including the required source, an omission annotated as the root cause. At step 22, it opens a \emph{Human Origins} video whose displayed content does not provide the requested measurement. Despite other local score peaks, ReCast ranks step 4 first, prioritizing the earlier annotated omission over the later browsing event.

\textbf{HC-42, premature completion (Figure~\ref{fig:hc-ag-case-study}(c)).}
The task asks which word was deleted in a federal rule amendment documented on Cornell Law School's website. Browsing the Federal Rules index at step 16 and the Witnesses article at step 24 produces local score peaks. At step 29, however, the Orchestrator declares the task complete while acknowledging that the retrieved page does not identify the requested word. This premature completion is the annotated root cause. ReCast assigns its highest score to step 29, ranking the unsupported completion decision above the earlier browsing steps.

\textbf{AG-116, substituting simulated data (Figure~\ref{fig:hc-ag-case-study}(d)).}
The task asks for the lowest sale price of a single-family house in Queen Anne in January 2023. With the transaction file unavailable, step 3 proposes analysis using an assumed CSV path and receives an elevated score. At step 7, the agent constructs simulated transactions instead of obtaining the actual records; the dataset annotates this substitution as the root cause. ReCast assigns its highest score to step 7, ranking the use of simulated data above the earlier analysis based on an assumed file path.

\textbf{AG-109, incorrect distance conversion (Figure~\ref{fig:hc-ag-case-study}(e)).}
The task seeks supermarkets within two blocks of Lincoln Park, Chicago, that offer ready-to-eat salads for under \$15. At step 1, the agent converts miles to blocks using an assumed rate of 20 blocks per mile, which the dataset identifies as incorrect for Chicago. A later discussion of candidate stores at step 5 also produces a local score peak. ReCast nevertheless ranks step 1 highest, selecting the annotated conversion error rather than the subsequent store recommendations.

\textbf{AG-120, unverified restaurant claims (Figure~\ref{fig:hc-ag-case-study}(f)).}
AG-120 addresses the same restaurant-search task as HC-50 through a different execution. At step 7, a missing Google Maps API key blocks automated verification, and the agent proposes manual checks. At step 8, another agent presents distances, dining options, and menu prices as verified without performing the required checks, which the dataset annotates as the root cause. ReCast's score rises at step 7 but peaks at step 8, prioritizing the unsupported claims over the preceding verification obstacle.

\clearpage

% !TEX root = ../../main.tex

\section{Prompt Templates}
\label{app:baseline-prompts}

\definecolor{PromptBlue}{RGB}{72,108,163}
\definecolor{PromptBackground}{RGB}{246,248,252}
\definecolor{PromptVariable}{RGB}{190,45,45}
\definecolor{PromptRole}{RGB}{64,82,112}
\newcommand{\promptvar}[1]{\textcolor{PromptVariable}{\{#1\}}}
\newcommand{\promptrole}[1]{%
  \textcolor{PromptRole}{\sffamily\bfseries #1}\par\vspace{0.2em}%
}
\newsavebox{\promptboxbody}
\newenvironment{prompttemplate}[1]{%
  \def\promptboxtitle{#1}%
  \setlength{\fboxsep}{7pt}%
  \begin{lrbox}{\promptboxbody}%
  \begin{minipage}{\dimexpr\linewidth-2\fboxsep-2\fboxrule\relax}%
  \vspace{0.15em}%
  \footnotesize\ttfamily\raggedright\sloppy
  \setlength{\parindent}{0pt}%
  \setlength{\parskip}{0.38em}%
}{%
  \vspace{0.05em}%
  \end{minipage}%
  \end{lrbox}%
  \par\medskip
  \noindent
  \setlength{\fboxsep}{5pt}%
  \colorbox{PromptBlue}{%
    \parbox{\dimexpr\linewidth-2\fboxsep\relax}{%
      \color{white}\sffamily\bfseries\small\strut\promptboxtitle
    }%
  }%
  \par\nobreak\vspace{-0.4pt}\nobreak
  \noindent
  \setlength{\fboxsep}{7pt}%
  \fcolorbox{PromptBlue}{PromptBackground}{\usebox{\promptboxbody}}%
  \par\medskip
}

We directly use the Who\&When implementation~\citep{zhang2025whowhen} and present its prompt templates below. The red fields denote trajectory-specific values inserted at inference time. The \promptvar{ground\_truth} field contains the benchmark-provided task answer, not the root-cause annotation. Line wrapping below is typographical and does not change the prompt content.

\subsection{All-at-Once Prompt}
\label{app:all-at-once-prompt}

All-at-Once presents the complete trajectory in one request and asks the model to identify the responsible agent and the first erroneous step.

\begin{prompttemplate}{All-at-Once Evaluation Template}
\promptrole{SYSTEM MESSAGE}
You are an AI assistant skilled in analyzing conversations.

\vspace{0.4em}
\promptrole{USER MESSAGE}
You are an AI assistant tasked with analyzing a multi-agent conversation history when solving a real world problem. The problem is: \promptvar{problem}

The Answer for the problem is: \promptvar{ground\_truth}

Identify which agent made an error, at which step, and explain the reason for the error. Here's the conversation:

\promptvar{conversation}

Based on this conversation, please predict the following:

1. The name of the agent who made a mistake that should be directly responsible for the wrong solution to the real world problem. If there are no agents that make obvious mistakes, decide one single agent in your mind. Directly output the name of the Expert.

2. In which step the mistake agent first made mistake. For example, in a conversation structured as follows:

\{\newline
\hspace*{1em}"agent a": "xx",\newline
\hspace*{1em}"agent b": "xxxx",\newline
\hspace*{1em}"agent c": "xxxxx",\newline
\hspace*{1em}"agent a": "xxxxxxx"\newline
\},

each entry represents a `step' where an agent provides input. The `x' symbolizes the speech of each agent. If the mistake is in agent c's speech, the step number is 2. If the second speech by `agent a' contains the mistake, the step number is 3, and so on. Please determine the step number where the first mistake occurred.

3. The reason for your prediction.

Please answer in the format:

Agent Name: (Your prediction)\newline
Step Number: (Your prediction)\newline
Reason for Mistake:
\end{prompttemplate}

\subsection{Step-by-Step Prompt}
\label{app:step-by-step-prompt}

Step-by-Step evaluates growing trajectory prefixes. At each step, the model receives the history up to and including that step. Evaluation terminates at the first response parsed as ``Yes''; otherwise, it advances to the next prefix.

\begin{prompttemplate}{Step-by-Step Evaluation Template}
\promptrole{SYSTEM MESSAGE}
You are a precise step-by-step conversation evaluator.

\vspace{0.4em}
\promptrole{USER MESSAGE AT STEP \promptvar{step}}
You are an AI assistant tasked with evaluating the correctness of each step in an ongoing multi-agent conversation aimed at solving a real-world problem. The problem being addressed is: \promptvar{problem}. The Answer for the problem is: \promptvar{ground\_truth}

Here is the conversation history up to the current step:

\promptvar{conversation}

The most recent step (\promptvar{step}) was by '\promptvar{agent}'.

Your task is to determine whether this most recent agent's action (Step \promptvar{step}) contains an error that could hinder the problem-solving process or lead to an incorrect solution. Please respond with 'Yes' or 'No' and provide a clear explanation for your judgment. Note: Please avoid being overly critical in your evaluation. Focus on errors that clearly derail the process.

Respond ONLY in the format:

1. Yes/No.\newline
2. Reason: [Your explanation here]
\end{prompttemplate}

\end{document}